\documentclass[11pt]{article}

\usepackage[preprint]{acl}

\usepackage{times}
\usepackage{latexsym}

\usepackage[T1]{fontenc}

\usepackage[utf8]{inputenc}
\usepackage{wrapfig}

\usepackage{microtype}

\usepackage{inconsolata}

\usepackage{graphicx}
\usepackage{subcaption}
\usepackage{hyperref}
\usepackage{url}
\usepackage{multirow}
\usepackage{wrapfig}
\usepackage{array}
\usepackage{caption}
\usepackage{arydshln} 
\usepackage{booktabs}
\usepackage{xcolor,colortbl}
\definecolor{lightgray}{rgb}{0.9,0.9,0.9}
\definecolor{mygray}{gray}{.9} 
\usepackage{amsmath}    
\usepackage{mathrsfs}   
\usepackage{xcolor}  
\usepackage{colortbl}  
\usepackage{textcomp}
\usepackage{booktabs}
\usepackage{arydshln}
\usepackage{booktabs}
\usepackage[breakable]{tcolorbox}
\usepackage{fontawesome5}
\definecolor{darkblue}{rgb}{0.0, 0.17, 0.58}
\definecolor{darkgreen}{rgb}{0.0, 0.5, 0.0}
 \usepackage{amssymb}
\title{ChunkLLM: A Lightweight Pluggable Framework for Accelerating LLMs Inference}

\author{Haojie Ouyang$^{1,}\thanks{\hspace{2pt}Equal contribution.}$, Jianwei Lv$^{2,}$\footnotemark[1], Lei Ren$^{2}$, Chen Wei$^{2}$, Xiaojie Wang$^{1,}$\thanks{\hspace{2pt}Corresponding author.}, Fangxiang Feng$^{1,}$\footnotemark[2]  \\
$^{1}$School of Artificial Intelligence, Beijing University of Posts and Telecommunications\\
$^{2}$Li Auto\\
\texttt{ouyanghaojie@bupt.edu.cn} 
}

\begin{document}
\maketitle

\begin{abstract}
Transformer-based large models 
suffer from severe computational inefficiency due to the quadratic complexity of self-attention with respect to input tokens.
Recently, block selection and compression methods have been proposed 
to alleviate this problem, but they either compromise semantic completeness or are inefficient in both training and inference.
To address these challenges, we propose ChunkLLM, a lightweight and pluggable framework. Its core consists of two novel components: the QK Adapter (comprising Q- and K-Adapters) and the Chunk Adapter. The QK Adapter is integrated into each Transformer layer. It performs dual functions: compressing features and acquiring chunk attention. The Chunk Adapter is placed at the bottommost layer of the model. It identifies chunk boundaries by leveraging contextual information. 
Notably, we design an attention distillation method to train the QK Adapter, which enhances the recall rate of key chunks. During inference, chunk selection is triggered only when the current token is identified as a chunk boundary, thereby accelerating the model.
Experimental evaluations show that
ChunkLLM achieves comparable performance on short-text benchmarks, while retaining 99.49\% of the performance on long-context benchmarks with only 52.10\% of key-value cache. Moreover, it is up to 4.48× faster than the vanilla Transformer when processing 120K-long texts.
\end{abstract}

\section{Introduction}

Transformer-based large models \citep{vaswani2017attention} have demonstrated exceptional performance across a diverse range of tasks, including natural language processing \citep{srivastava2025instruction,zhang2024event} and computer vision \citep{jiang2025wise}. However, they have also faced significant challenges in terms of computational efficiency, particularly when scaling to larger structures and large context inputs. A core issue of efficiency limitations lies in the self-attention module, whose computational complexity is a quadratic relationship with respect to the number of input tokens. Such deficiencies in computational efficiency exert a profound impact on both the training complexity and inference latency of large models.

Efficiency optimization of Transformer has emerged as a pivotal research domain, with efforts predominantly converging into three methodological paradigms.
\textbf{Linear attention}, such as Mamba \citep{dao2024transformers}, RWKV \citep{peng2023rwkv,peng2024eagle}, and RetNet \citep{sun2023retentive}, seek to approximate and substitute the traditional softmax-based self-attention mechanism. 
However, the fundamental architectural disparities between linear attention and conventional attention mechanisms introduce non-trivial challenges: adapting pre-existing Transformer models to integrate linear attention often incurs prohibitive conversion costs\citep{mercat2024linearizing,wang2024mamba,bick2024transformers}, while alternative strategies necessitate end-to-end training of entirely new model from scratch\citep{li2025minimax}.
Another optimization paradigm is \textbf{Sparse attention}, which leverages predefined structural constraints, such as sink-based attention mechanisms \citep{DBLP:conf/iclr/XiaoTCHL24} or sliding window attention mechanisms \citep{beltagy2020longformer}, to exploit this sparsity. While these methods may yield certain effects, they often rely heavily on specific tasks, which can limit the overall generalization ability of the model. Dynamic sparse attention mechanisms \citep{tang2024quest,jiang2024minference,liu2024retrievalattention} filter out subsets of tokens during the inference phase. Although such methods can reduce the computational load of long sequences, they fail to significantly lower the high training costs of long-context models, making it difficult for large language models to efficiently scale to context-processing tasks with million-level token sizes. 
\textbf{Chunk Selective attention}, a special type of sparse attention, can be primarily categorized into two paradigms: fixed chunk \citep{DBLP:journals/corr/abs-2502-13189,DBLP:conf/acl/YuanGD0ZZXWW0WR25,DBLP:journals/corr/abs-2502-14477} and separators-based dynamic chunk \citep{DBLP:journals/corr/abs-2412-12094}. Both approaches partition the input into discrete chunks: the former conducts partitioning with a fixed length, which gives rise to semantic incompleteness; the latter utilizes separators for partitioning, yet ambiguities often emerge. For example, periods frequently occur in numerical values or abbreviations. Furthermore, during the inference phase, these methods necessitate chunk selection for each generated token, incurring additional computational overhead. It is thus evident that existing efficient approaches still exhibit inherent limitations.

To address the aforementioned challenges, We propose ChunkLLM, which can be directly constructed by integrating two lightweight and trainable modules into existing LLMs: \textbf{QK Adapter} and \textbf{Chunk Adapter}. The Chunk Adapter connects to the output of the bottommost Transformer layer and used for identify if a token is the last token of a chunk. The QK Adapter is in parallel with Q and K matrix at each Transformer layer. It maps full attention scores to chunk attention scores, and trained by a distillation approach. 

The QK Adapter fulfills feature compression and the generation of chunk attention scores. To train the QK Adapter, we propose an attention distillation approach designed to enhance the recall rate of key chunks. During training, LLM parameters are kept frozen, with the Kullback–Leibler (KL) divergence between chunk attention scores and full attention scores serving as a guidance signal for optimization. The Chunk Adapter determines whether a token corresponds to a chunk boundary by leveraging contextual semantic information. During the inference phase, we exploit the Intra-Chunk Attention Consistency (ICAC) pattern such that chunk selection is only updated when the current token is identified as a chunk boundary, which substantially enhances inference efficiency. Furthermore, ChunkLLM can achieve inference performance comparable to that of models optimized for 120K context lengths, despite being trained solely on 4K context lengths, thereby substantially reducing the training overhead associated with 120K context scaling. Experimental results validate that ChunkLLM yields a 4.48× speedup relative to the vanilla Transformer when processing 120K long texts.

Our contributions are summarized as follows:

\begin{itemize} 
    \item We introduce ChunkLLM, which integrates two lightweight and pluggable components ---the QK Adapter and the Chunk Adapter--- into existing LLMs. It only requires fine-tuning these components based on the original model architecture. 

    \item We propose an attention distillation-based training approach for the QK Adapter, which leverages KL divergence to align chunk attention with full attention, thereby enhancing the recall rate of key chunks. Furthermore, we introduce a novel ICAC pattern, which yields notable improvements in inference efficiency for long-context scenarios.

    \item Experimental evaluations show that, compared to the vanilla Transformer, ChunkLLM 
    maintains 99.49\% of the performance on long-context benchmarks  while using only 52.10\% of the key-value cache (KVcache). 
    This reduction in memory footprint contributes to a substantial speedup, reaching up to 4.48× on 120K-long texts.
    \end{itemize}

\section{Method}

\begin{figure}
\centering
\includegraphics[width=0.85\linewidth]{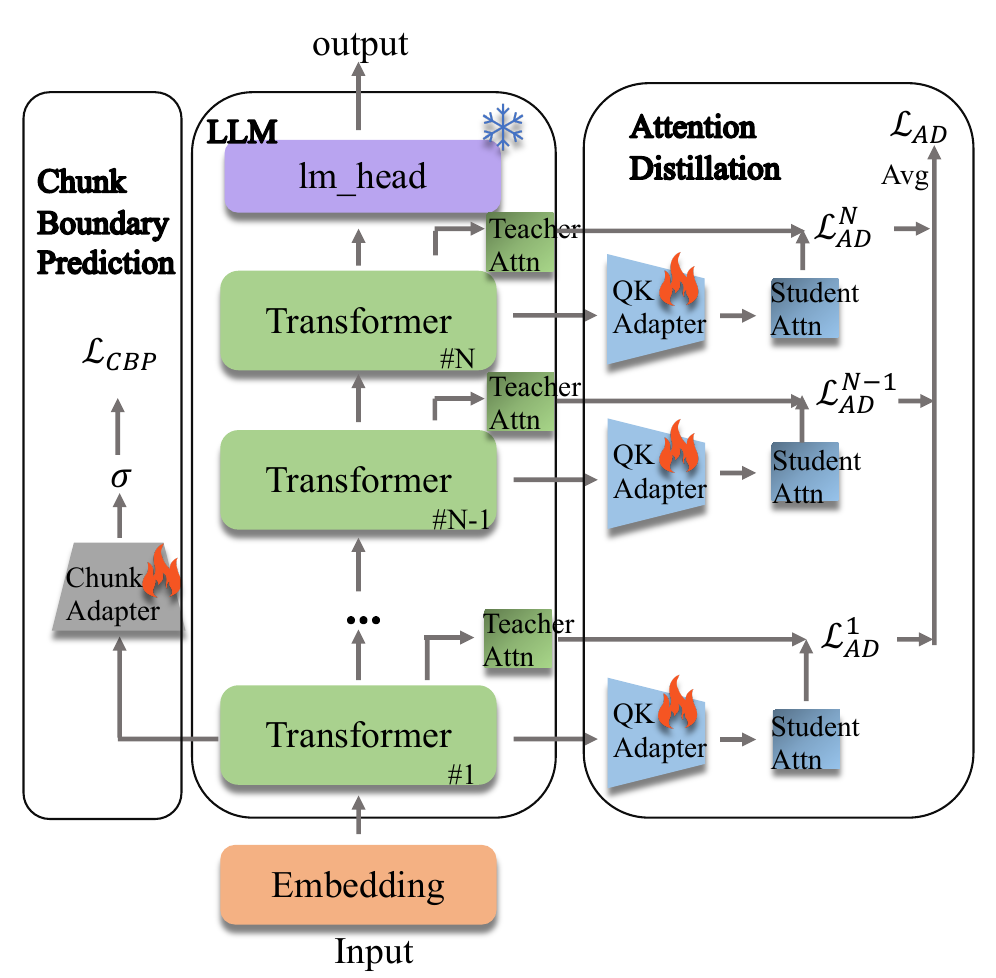}
\caption{The framework of ChunkLLM. 
}
\label{fig:train_model}
\end{figure}

The framework of ChunkLLM is shown in Figure \ref{fig:train_model}. ChunkLLM can be built on any existing transformer-based LLMs. Two extra lightweight and pluggable modules are designed to support chunk-related capability. One is Chunk Adapter, which is used to identify chunk boundaries. The other is the Q Adapter and K Adapter, which is tailored for efficient feature compression and chunk selection. This section elaborates on the details of the two modules.

\subsection{Chunk Adapter}


In existing research on chunk-sparse attention, most methods adopt fixed-size chunking. However, this approach fails to account for the semantic structure of the sequence, leading to context fragmentation and incomplete retrieval of key information. To address this issue, we employ a sentence segmentation tool to preprocess the data and train a semantically aware Chunk Adapter.

The Chunk Adapter is a one-layer forward neural network (FNN) classifier for chunk boundary prediction. Its input is the output of the first layer of the LLM, and the output is if or not the token is a chunk boundary, as depicted in the Figure \ref{fig:train_model}.

For an input $\mathrm{X}=\{x_{1},x_{2},...,x_{n-1},x_{n}\}$ with $n$ tokens and their corresponding labels $\mathrm{Y}=\{y_{1},y_{2},...,y_{n-1},y_{n}\}$, $y_{i} \in \{0,1\}$ , where 1 indicates that the token $x_{i}$ is a chunk boundary, 0 indicates that it does not, $\mathrm{\textbf{H}}_i^{l_{1}}$ be the output of $x_{i}$ at first layer.
The FNN based Chunk Adapter is given as in equation ~\ref{eq:chunk-label} 

\begin{equation}
    \hat{y}_{i} = \begin{cases}
    1, & \mathrm{Sigmoid}(\mathrm{FFN}(\textbf{H}_i^{l_{1}})) > \alpha, \\
    0, & otherwise
    \end{cases}
    \label{eq:chunk-label}
\end{equation}

For training the chunk adapter, we employ the binary cross-entropy loss (BCE) (equation ~\ref{eq:bce-loss}) as the objective function. Detailed information on the training dataset will be given in the experiment part. 

\begin{equation}
    \mathcal{L}_{CBP} = -\frac{1}{n}\sum_{i=1}^{n}[y_{i}\cdot log(\hat{y}_{i}) + (1-y_{i})\cdot log(1-\hat{y}_{i})]
    \label{eq:bce-loss}
\end{equation}

\subsection{QK Adapter}

At each layer of the LLM, we incorporate a Q-Adapter and a K-Adapter which used to compress the attention and select the chunks. 

For each layer, let $\textbf{Q}$ and $\textbf{K}$ be the attention matrix respectively, $c$ be the chunk number of the input, $Index\_c = \{i_1,i_2,...i_c\}$ is the index set of chunk boundary tokens. Let $\hat{\textbf{K}}$ be the K matrix of these tokens. We then calculate chunk attention scores as follows: 

\begin{equation}
\begin{aligned}
    &\textbf{A}^{s} = Softmax(\frac{Mul(\bar{\textbf{Q}},\bar{\textbf{K}}^{T})}{\sqrt{d_{k}}}) \\ 
    & \bar{\textbf{Q}} = FFN_{Q}(\textbf{Q}), \bar{\textbf{K}} = FFN_{K}(\hat{\textbf{K}}) 
\end{aligned}
\label{eq:a_s attn}
\end{equation}
where $FFN_{Q}$ and $FFN_{K}$ is proposed Q-Adapter and K-Adapter, respectively, $\bar{\textbf{Q}} \in \mathbb{R}^{n\times d_{k}}, \bar{\textbf{K}} \in \mathbb{R}^{c\times d_{k}}$, $d$ is the dimension of the model, and $d_k$ is the dimension of head. $d_{k} \ll d$.

\textbf{Attention Distillation} 
We propose an attention distillation strategy to train the Q-Adapter and K-Adapter. Where, we treat $\textbf{A}^{s}$ as student attention, and a type of aggregation of original attention $\textbf{A}^{t}$ which is given in follow as teacher attention. The objective is to align the student’s chunk attention with that of the teacher, improving the recall performance for key chunks. As shown in Figure \ref{fig:train_model}. 

For the sake of descriptive simplicity, we use a single head as an illustrative example to show how to aggregate original attention. The calculation procedure is detailed as follows:

\begin{equation}
\begin{aligned}
    &\textbf{A}^{t} = Aggregate(\textbf{A}) \\
    &\textbf{A} = Softmax(\frac{Mul(\textbf{Q},\textbf{K}^{T})}{\sqrt{d_{k}}}) \qquad \textbf{A} \in \mathbb{R}^{n\times n}
\end{aligned}
    \label{eq:att-teacher}
\end{equation}

where $\textbf{Q}\in \mathbb{R}^{n\times d_{k}}$ and $\textbf{K}\in \mathbb{R}^{n\times d_{k}}$ are the matrices of query and key for one attention layer. For brevity, the mask operation is omitted from the description. 

$Aggregate$ denotes the operation of summing the token scores within a single chunk. 
Specifically, for the matrix \textbf{A}, the attention scores of tokens within each chunk are accumulated along the column direction, representing the attention scores of tokens to the chunk.
For multi-head attention, we compute the average along the head dimension, yielding matrix $\textbf{A}^{t}$.

We employ the Kullback-Leibler (KL) divergence as the loss function for attention distillation to guide the student model $\textbf{A}^{s}$ in approximating the teacher model’s attention scores $\textbf{A}^{t}$ :
\begin{equation}
    \mathcal{L}_{AD}^{N} = KL(\textbf{A}^{t}||\textbf{A}^{s})
\end{equation}

We average the KL divergence losses across the N layers to obtain the final attention distillation loss:
\begin{equation}
    \mathcal{L}_{AD} = \frac{1}{N}\sum_{i}^{N}\mathcal{L}_{AD}^{i}
\end{equation}
During the training phase, the parameters of the backbone network are frozen, with only the Chunk Adapter and QK Adapter undergoing training, thereby achieving efficient training.

\subsection{Inference}

\begin{figure}[t]
\centering
\includegraphics[width=1.06\linewidth]{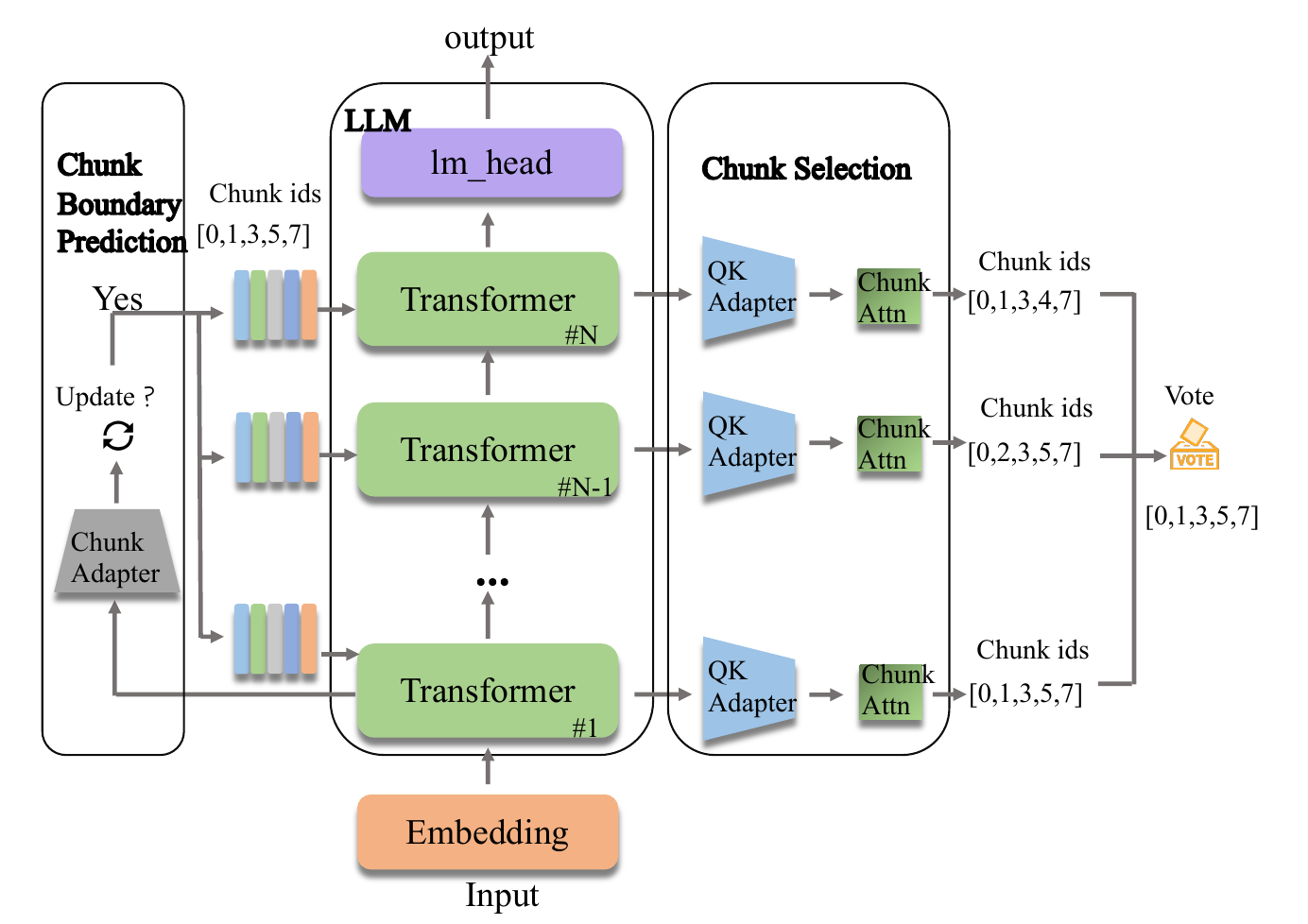}
\caption{The decoding process of ChunkLLM.}
\label{fig:inference_model}
\end{figure}


We use sparse attention in the decoding process during the inference phase. The decoding process of ChunkLLM is depicted in Figure \ref{fig:inference_model}, encompassing two primary steps: chunk boundary prediction and chunk selection.

\textbf{Chunk Boundary Prediction} During inference, the Chunk Adapter performs binary classification on each input token to identify the positions of chunk boundary tokens, which are then used for Chunk Selection.

\textbf{Chunk Selection} In ChunkLLM, two separate KV-caches are maintained, which we refer to as the Cold KV-cache and the Hot KV-cache. The former stores the KV vectors of all tokens and is kept in either GPU memory or CPU memory. The latter stores the KV vectors of the chunks retrieved during the chunk selection stage and is actually used in the attention computation.
In this stage, given an input token, the corresponding q, k, v vectors are computed. The k and v vectors are stored in the Cold KV-cache. Then, referring to equation ~\ref{eq:a_s attn}, $\textbf{A}^{s}$ is computed, and the chunks corresponding to the top-k scores are selected. The k and v vectors of these chunks are loaded into the Hot KV-cache, where they are used together with the q vector to compute the actual attention scores, producing the predicted token.

\begin{figure}[t]
\centering
\includegraphics[width=\linewidth]
{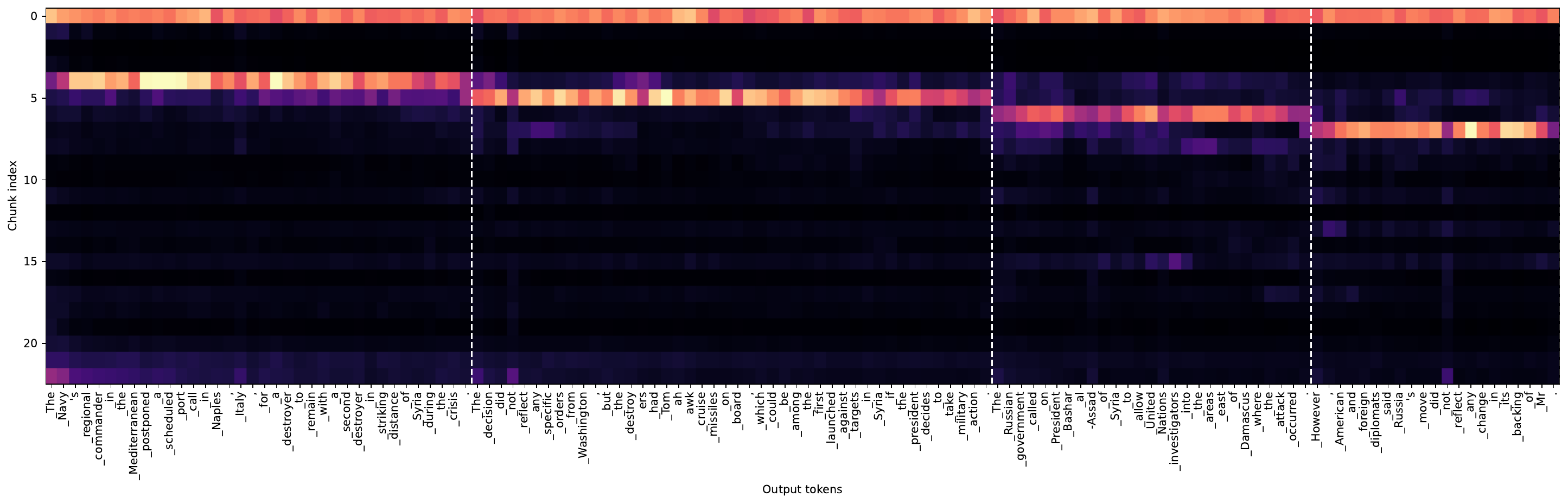}
\caption{Attention visualization of chunk selection during the inference phase. 
The test case is from the LongBench summarization task.
}
\label{fig:icac}
\end{figure}

\textbf{ICAC} 
\label{ICAC_example}
We find a phenomenon during the model inference, as illustrated in Figure \ref{fig:icac}. The chunks attended to by tokens within a generated chunk exhibit substantial consistency, whereas chunk updates predominantly occur at chunk boundaries. We name this phenomenon the "\textbf{I}ntra-\textbf{C}hunk \textbf{A}ttention \textbf{C}onsistency (ICAC)".

ICAC makes it possible to save computational cost in chunk selection. Specifically, the update of the Cold KV-cache by Chunk Selection is only required when the Chunk Adapter identifies a new chunk boundary token; at all other times, the Cold KV-cache can be reused.


\textbf{Vote} In our experiments, we observed that not all layers of large language models exhibit sparse attention characteristics, and applying sparse attention distillation to these layers fails to achieve the expected results. We illustrate this phenomenon in Figure \ref{fig:top_k_chunks}. One solution is to fall back to full attention implementation. However, we found that, considering the results across all layers, the retrieved chunks are highly consistent with the key chunks. Therefore, we propose addressing this issue by aggregating the retrieval results from all layers, counting votes to select the top-k chunks with the highest votes as the shared chunks for all layers.

\section{Experiments and Results}

\subsection{Experimental Settings}

\subsubsection{Model and Baselines}

Two representative open-source models, Qwen2.5-7B \citep{qwen2.5} and Llama3.1-8B \citep{dubey2024llama}, are chosen as the target models for evaluation. We select StreamingLLM \citep{DBLP:conf/iclr/XiaoTCHL24}, SepLLM \citep{DBLP:journals/corr/abs-2412-12094} and SnapKV \citep{DBLP:conf/nips/LiHYVLYCLC24} as the baselines to benchmark the proposed method. 
Detailed settings of the experimental parameters and training datasets are provided in Appendix \ref{paramer_set}.

\begin{table*}[h]
\setlength\tabcolsep{3pt} 
	\centering
    \scriptsize
	\begin{tabular}{llcccccccccccccccc}
	    \hline
	    \multirow{2}{*}{Methods} & \multicolumn{3}{c}{SDQA} & \multicolumn{3}{c}{MDQA} & \multicolumn{2}{c}{Summary} & \multicolumn{3}{c}{Few-shot} & \multicolumn{2}{c}{Synthetic} & \multicolumn{2}{c}{Code} & \multirow{2}{*}{Avg} & \multirow{2}{*}{KV}\\
        \cmidrule(lr){2-4} \cmidrule(lr){5-7} \cmidrule(lr){8-9}\cmidrule(lr){10-12} \cmidrule(lr){13-14} \cmidrule(lr){15-16}
        & NQA & Qasper & MFQA & HQA & Musi & 2WQA & GR & QMS & SAM & TQA & TREC & PC & PR-en & LCC & RB-P & & \\
        \hline
        Qwen2.5-7B & 33.76 & 51.40 & 57.17 & 56.76 & 29.40 & 38.09 & 31.57 & 24.40 & 46.73 & 92.50 & 72.00 & 2.00 & 66.50 & 62.86 &53.53 &  47.91 & 100.00\\ 
        \cdashline{1-18}[2pt/1pt]
        ChunkLLM & 31.09 & 50.52 & \textbf{56.95} & 55.61 & \textbf{29.41} & \textbf{38.27} & \textbf{31.25} & 23.16 & 46.53 & \textbf{92.50} & \textbf{72.50} & \textbf{3.00} & \textbf{68.00} & 62.72 & \textbf{53.50} & \textbf{47.67} & 52.10\\
        SepLLM & 26.43 & 50.18 & 50.29 & 47.25 & 22.83 & 36.34 & 28.31  & 21.14 & 46.71 & 91.50 & 71.50& 1.50 & 65.50 & \textbf{62.88} & 52.84 & 45.14 & 51.85\\
        StrmLLM & 28.88 & 50.37 & 50.47 & 51.60 & 24.55 & 37.76 & 30.36 & 22.27 & \textbf{46.84} & 91.50 & 72.00 & 2.50 & 67.50 & 62.68 & 53.07 & 46.16 & 67.75 \\
        SnapKV & \textbf{31.68} & \textbf{50.56} & 56.26 & \textbf{56.11} & 28.19 & 37.64 & 28.95  & \textbf{24.51} & 46.44 & 92.50 & 72.00 & 1.50 & 66.00 & 62.36 & 53.47 & 47.21 & 51.62\\
        \hline
        Llama3.1-8B & 40.88 & 51.72 & 58.47 & 52.68 & 35.80 & 42.52 & 30.64 & 24.44 & 47.10 & 92.83 & 73.00 & 0.50 & 39.00 & 68.56 & 67.48 &  48.37 & 100.00\\
        \cdashline{1-18}[2pt/1pt]
        ChunkLLM & 39.14 & 49.93 & 53.45 & 52.29 & 34.40 & \textbf{42.98} & \textbf{31.05} &  23.89 &   \textbf{47.20} & \textbf{93.08} & 72.50 & \textbf{0.50} & 38.50 & 68.31 & 66.99 & \textbf{47.62} & 53.85\\
        SepLLM & 36.23 & \textbf{51.16} & 51.81 & 47.70 & 27.83 & 40.85 & 27.12 &  21.84 & 46.89 & 92.83 & 72.00 & 0.50 & 37.50 & 68.50 & 66.54 & 45.95 & 53.08\\
        StrmLLM & 35.94 & 50.96 & \textbf{54.13} & 50.54 & 30.95 & 41.83 & 27.50 &  23.06 & 46.23 & 92.83 & \textbf{73.00} & 0.50 & \textbf{39.00} & 68.43 & \textbf{67.00} & 46.80 & 68.68 \\
        SnapKV & \textbf{40.17} & 51.16 & 55.73 & \textbf{52.83} & \textbf{35.52} & 42.78 & 28.21 & \textbf{24.31} & 45.60 & 92.88 & 72.00 & 0.50 & 32.66 & \textbf{68.64} & 66.86 & 47.32 & 53.75 \\
	    \hline

	\end{tabular}
	\caption{Experimental results on LongBench.}
	\label{tab:longbench}
\end{table*}

\subsubsection{Benchmarks}

\textbf{Long Context Benchmarks} We select two long-context evaluation datasets, LongBench \citep{DBLP:conf/acl/BaiLZL0HDLZHDTL24} and Needle In A Haystack (NIAH) \citep{niah}, to assess the model's long-context ability.
For detailed data and description, see Appendix \ref{append_longbench}.


\textbf{Short Context Benchmarks} The selection of the evaluation datasets is primarily centered on the model's performance in three key dimensions: \textbf{General Knowledge}, which evaluates the model's breadth of knowledge coverage and the accuracy of its knowledge, MMLU \citep{DBLP:conf/iclr/HendrycksBBZMSS21} (5-shot), SciQ \citep{DBLP:conf/aclnut/WelblLG17} (5-shot), OpenBookQA \citep{DBLP:conf/emnlp/MihaylovCKS18} (25-shot); \textbf{Question Answering}, which evaluates the model's capabilities in question understanding and information matching, CommonsenseQA \citep{DBLP:conf/naacl/TalmorHLB19} (5-shot), Social IQA \citep{DBLP:conf/emnlp/SapRCBC19} (15-shot), PIQA \citep{bisk2020piqa} (25-shot); and \textbf{Reasoning} which evaluates the model's capabilities in logical abstraction and complex decision-making, HellaSwag \citep{DBLP:conf/acl/ZellersHBFC19} (10-shot), WinoGrande \citep{DBLP:conf/aaai/SakaguchiBBC20} (25-shot), ARC-c/ARC-e \citep{clark2018think} (25-shot).

\subsection{Main Results}

\subsubsection{Results on LongBench} 

We set the top-k ratio to 45\% and the number of local chunks to 15 for ChunkLLM. 
The experimental results on the LongBench using Qwen2.5-7B and Llama3.1-8B are presented in Table \ref{tab:longbench}.
(The sensitivity experiments of hyperparameters top-k rate and local chunks are elaborated upon in Appendix \ref{top-k} and \ref{local-chunks}, respectively.) Here, "StrmLLM" represents StreamingLLM \citep{DBLP:conf/iclr/XiaoTCHL24}. 

We take Qwen2.5-7B as an example for analysis, and the same conclusion holds for Llama3.1-8B. The following observations can be made: (1) In terms of overall average performance, ChunkLLM attains the optimal performance when compared to SepLLM, StreamingLLM and SnapKV, with respective improvements of 2.51, 1.07 and 0.46 (47.67 v.s. 45.14 v.s. 46.16 v.s. 47.21). In contrast to the short-text benchmark in Subsection \ref{shot_result}, ChunkLLM demonstrates a remarkable improvement in long-context evaluation, which validates the advantage of ChunkLLM in retrieving key chunk information during long-context reasoning.
(2) Notably, in the MDQA task, ChunkLLM yields a substantial improvement over SepLLM. We argue that the core challenge of MDQA lies in the dispersion of critical information across distinct positions within the context, which places high demands on the model’s context comprehension capability.
SepLLM leverages separators as chunk features, which is plagued by constrained representational capacity and the problem of chunk semantic incompleteness. By contrast, ChunkLLM enriches the representational capacity of chunks via attention distillation, which enhances the recall rates of critical chunks. This, in turn, effectively boosts the model’s long-context understanding capability.
(3) Given a comparable level of KV cache usage, ChunkLLM attains 99.49\% of the vanilla model’s performance, relative to SepLLM, StreamingLLM and SnapKV, findings that further substantiate the superiority of ChunkLLM in long-context scenarios.

\subsubsection{Results on NIAH}

\begin{figure*}[hbt]
    \centering
    \includegraphics[scale=0.4]{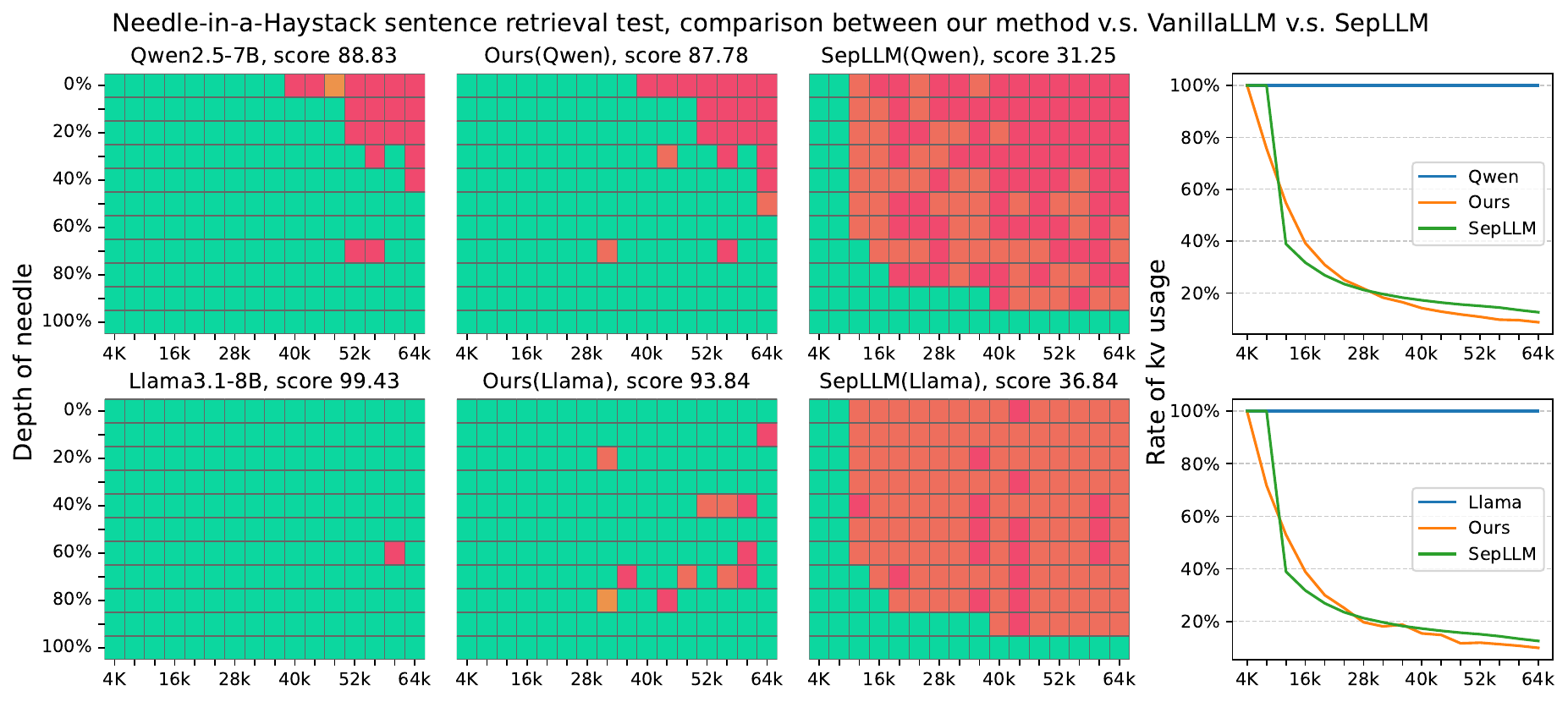}
    \caption{Needle-in-a-Haystack retrieval accuracy across context positions with 64k context length. The last column represents the KV-cache utilization rate.
    }
    \label{fig:NIAH}
\end{figure*}

\begin{figure*}[hbt]
    \centering
    \includegraphics[scale=0.4]{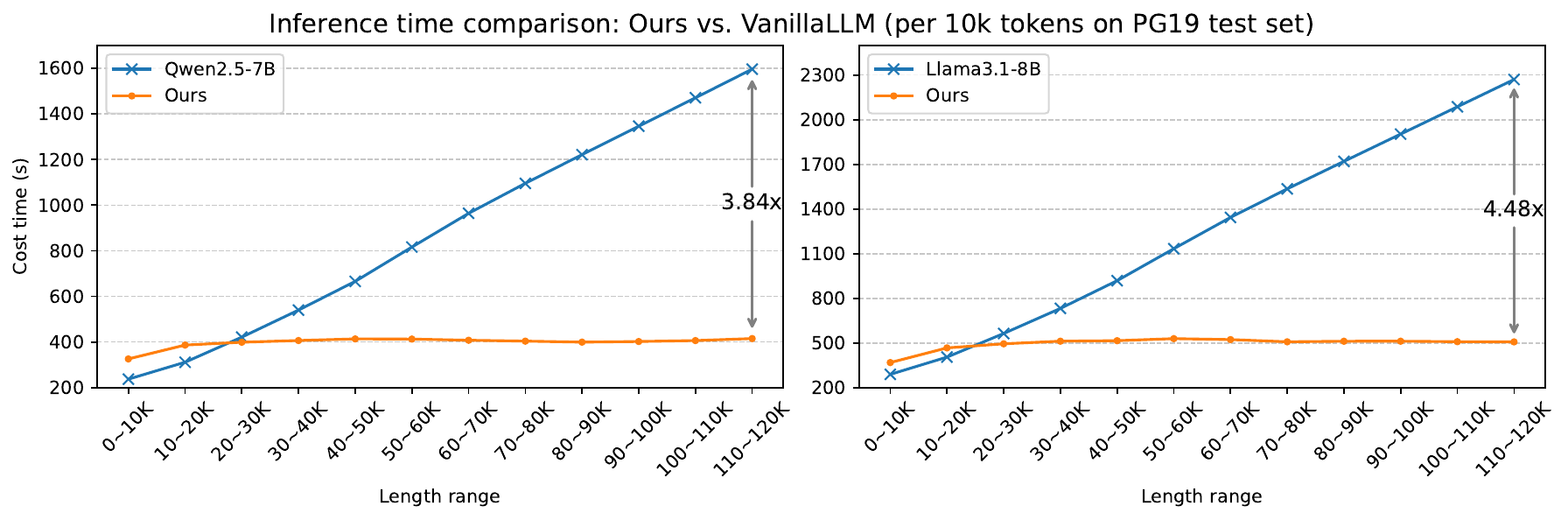}
    \caption{Comparison of inference time per 10k tokens in the generation process on PG19 test set.
    }
    \label{fig:enfficiency}
\end{figure*}

We set the top-k to 256 and the number of local chunks to 16 for ChunkLLM.
As depicted in Figure \ref{fig:NIAH}, ChunkLLM outperforms SepLLM across all scenarios in the 64K-context NIAH evaluation conducted on Qwen2.5-7B and Llama3.1-8B, achieving superior performance. Notably, in scenarios where the context length exceeds 12K, SepLLM exhibits near-total loss of retrieval capability (visualized in red), whereas ChunkLLM retains performance comparable to the vanilla model. This discrepancy is primarily attributed to ChunkLLM's attention distillation mechanism, which strengthens the feature representational capacity of chunks. Consequently, during chunk selection, the model effectively identifies critical chunks with higher query relevance, leading to improved inference performance.
Additionally, ChunkLLM exhibits a reduced KV-Cache utilization rate relative to SepLLM, which further corroborates the effectiveness of key chunk retrieval.
We also conduct experiments with StreamingLLM, as shown in Appendix \ref{NIAH_stream}.

\subsubsection{Inference Efficiency}

We conduct runtime evaluations of Vanilla and ChunkLLM for 120K-token generation tasks on the PG19\citep{rae2019compressive} test set, with metrics recorded every 10K tokens. 
We set the top-k to 256 and the number of local chunks to 16 for ChunkLLM.
As shown in Figure \ref{fig:enfficiency}, as the number of generated tokens increases, Vanilla’s inference time rises linearly, while ChunkLLM maintains persistent stability in time consumption. In the 110K–120K token generation phase, ChunkLLM outperforms Vanilla by speedups of 3.84× and 4.48×, which corroborates the efficacy of the proposed ICAC mechanism. During ChunkLLM’s inference phase, chunk updates occur exclusively at chunk boundaries, minimizing the updates frequency  and thereby boosting inference efficiency. 
The inference time of all sub-processes can be found in Appendix \ref{sub-process}.
We also conduct supplementary experiments using the FineWeb-Edu dataset, from which 1000 test corpora of 4k length are sampled. For the task of chunk boundary prediction, we evaluate its performance using three key metrics: precision, recall, and F1-score. The calculated results are 98.31, 95.54, and 96.91, respectively. Such promising performance indicators serve to verify the reliability and effectiveness of our chunk boundary prediction task.

\begin{figure}
\centering
\scriptsize
	\begin{tabular}{clcr}
	    \hline
	Length & Methods & PPL & Total Time(s)\\
        \hline
        \multirow{4}{*}{120K}& Qwen2.5-7B & 14.41 & 10,684.31\\ 
        \cdashline{2-4}[2pt/1pt]
        & ChunkLLM & 16.23 & 4,782.62\\
        \cline{2-4}
       
        & Llama3.1-8B & 11.93 & 14,906.19\\
        \cdashline{2-4}[2pt/1pt]
        & ChunkLLM & 12.89 & 5,963.83\\
	    \hline

	\end{tabular}

\captionof{table}{The perplexity and running time comparison on the PG19 test set.}

\label{fig:ppl}
\end{figure}

\begin{table*}[t]
	\centering
    \scriptsize
	\begin{tabular}{llccccccccccc}
	    \hline
	    \multirow{2}{*}{Methods} & \multicolumn{3}{c}{General Knowledge} & \multicolumn{3}{c}{Question Answering} & \multicolumn{4}{c}{Reasoning} & \multirow{2}{*}{Avg} & \multirow{2}{*}{KV}\\
        \cmidrule(lr){2-4} \cmidrule(lr){5-7} \cmidrule(lr){8-11}
        & MMLU & SciQ & OQA & CQA & SIQA & PIQA & Heag & WG & ARC-c & ARC-e &  &  \\
        \hline
        Qwen2.5-7B & 74.25 & 97.00 & 52.80 & 84.52 & 58.44 & 81.72 & 80.24 & 77.27 & 63.82 & 87.21 & 75.73 & 100.00\\ 
        \hdashline[2pt/1pt]
        ChunkLLM & 72.51 & 96.60 & \textbf{52.40} & \textbf{84.68} & 58.34 & \textbf{81.77} & 80.08 & \textbf{76.87} & \textbf{63.65} & \textbf{87.21} & 75.41 & 45.47\\
        SepLLM & 73.07 & \textbf{96.70} & 52.20 & 84.19 & 58.25 & 81.41 & 80.10 & 76.48 & 62.94 & 86.11 & 75.15 & 50.20\\
        StrmLLM & 73.31 & 96.60 & 52.00 & 84.28 & 58.19 & 81.39 & 79.76 & 76.64 & 62.79 & 86.36 & 75.13 & 45.14\\
        SnapKV & \textbf{74.05} & 96.60 & 52.40 & 84.03 & \textbf{58.40} & 81.61 & \textbf{80.18} & 76.71 & 63.28 & 87.16 & \textbf{75.44} & 45.57\\
        \hline
        Llama3.1-8B & 65.30 & 97.60 & 48.00 & 74.28 & 54.04 & 83.19 & 81.76 & 80.03 & 57.85 & 84.55 & 72.66 & 100.00\\
        \hdashline[2pt/1pt]
        ChunkLLM & \textbf{64.78} & 97.30 & \textbf{48.40} & \textbf{74.45} & \textbf{54.76} & 82.75 & \textbf{81.84} & 79.01 & 57.68 & 84.55 & \textbf{72.55} & 45.04\\
        SepLLM & 64.32 & \textbf{97.40} & 47.40 & 74.10 & 54.25 & \textbf{83.03} & 81.68 & 79.01 & \textbf{57.93} & 84.09 & 72.32 & 50.32 \\
        StrmLLM & 61.19 & 97.20 & 48.00 & 73.79 & 53.94 & 81.56 & 80.14 & 78.22 & 56.91 & 83.59 & 72.45 & 45.30\\
        SnapKV & 62.57 & 97.70 & 47.20 & 74.37 & 54.09 & 82.75 & 81.66 & \textbf{79.72} & 57.34 & \textbf{84.80} & 72.49 & 45.72\\
	    \hline

	\end{tabular}
	\caption{Experimental results on short context benchmarks.}
	\label{tab:merge_shottext}
\end{table*}

\begin{table*}[h]
\setlength\tabcolsep{3pt} 
	\centering
    \scriptsize
	\begin{tabular}{llccccccccccccccc}
	    \hline
	    \multirow{2}{*}{Methods} & \multicolumn{3}{c}{SDQA} & \multicolumn{3}{c}{MDQA} & \multicolumn{2}{c}{Summary} & \multicolumn{3}{c}{Few-shot} & \multicolumn{2}{c}{Synthetic} & \multicolumn{2}{c}{Code} & \multirow{2}{*}{Avg}\\
        \cmidrule(lr){2-4} \cmidrule(lr){5-7} \cmidrule(lr){8-9}\cmidrule(lr){10-12} \cmidrule(lr){13-14} \cmidrule(lr){15-16}
        & NQA & Qasper & MFQA & HQA & Musi & 2WQA & GR & QMS & SAM & TQA & TREC & PC & PR-en & LCC & RB-P &\\
        \hline
        Qwen2.5-7B & 33.76 & 51.40 & 57.17 & 56.76 & 29.40 & 38.09 & 31.57 & 24.40 & 46.73 & 92.50 & 72.00 & 2.00 & 66.50 & 62.86 &53.53 &  47.91\\ 
        \cdashline{1-17}[2pt/1pt]
        ChunkLLM & 31.09 & 50.52 & \textbf{56.95} & \textbf{55.61} & \textbf{29.41} & 38.27 & 31.25 & 23.16 & 46.53 & \textbf{92.50} & \textbf{72.50} & \textbf{3.00} & \textbf{68.00} & 62.72 & 53.50 & 47.67\\
         w/o vote & 30.78 & 45.00 & 51.53 & 54.96 & 28.00 & 38.17 & 30.82 & 23.49 & 46.58 & 90.50 & 70.00 & 2.50 & 67.50 & 62.54 & 52.78 & 46.34\\
         w/o ICAC &\textbf{31.36} & \textbf{50.94} & 56.84 & 55.61 & 28.50 & \textbf{38.59} & \textbf{31.86} & \textbf{23.93} & \textbf{46.66} & 92.50 & 72.50 & 2.50 & 67.50 & \textbf{63.16} & \textbf{53.88} & \textbf{47.76}\\
	    \hline

	\end{tabular}
	\caption{Ablation Study on LongBench, w/o vote: remove vote mechanism, w/o ICAC: remove ICAC pattern.}
	\label{tab:longbench_ablation}
\end{table*}

We conduct experiments where these models generated 120K tokens, evaluating both total inference time and average perplexity (ppl) on the PG19 test set, and results are summarized in Table \ref{fig:ppl}. Compared to the vanilla model, ChunkLLM yields a slight enhancement in ppl alongside a significant decrease in total inference time. The underlying reason is that while the vanilla model maintains semantic integrity, it incurs linearly increasing inference time as generated token count rises. Conversely, ChunkLLM reduces computational burden and speeds up inference by leveraging its chunk selection and ICAC mechanisms.

\subsubsection{Results on Short Text}
\label{shot_result}

The experimental results for short texts are presented in Table \ref{tab:merge_shottext}. The following conclusions can be drawn: (1) ChunkLLM performs comparably to or better than StreamingLLM, SepLLM, and SnapKV in terms of overall average metrics across both Qwen2.5-7B and Llama3.1-8B, achieving 99.57\% and 99.84\% of the Vanilla model's performance, respectively. 
Notably, ChunkLLM attains optimal performance across 5 out of the 10 evaluation tasks, validating its efficacy in short-text task scenarios.
(2) We perform statistical analyses on the average utilization rate of the KV cache. In comparison with SepLLM, ChunkLLM achieves superior performance while consuming a lower volume of KV cache (45.47\% v.s. 50.20\%). Specifically, on the Llama3.1-8B model, ChunkLLM not only exhibits the minimal KV cache usage but also outperforms both SepLLM and StreamingLLM in terms of performance metrics. This finding further validates the precision of ChunkLLM in chunk recall, achieving a balanced trade-off between performance and memory consumption.

\begin{figure}[t]
\centering
\includegraphics[width=\linewidth]{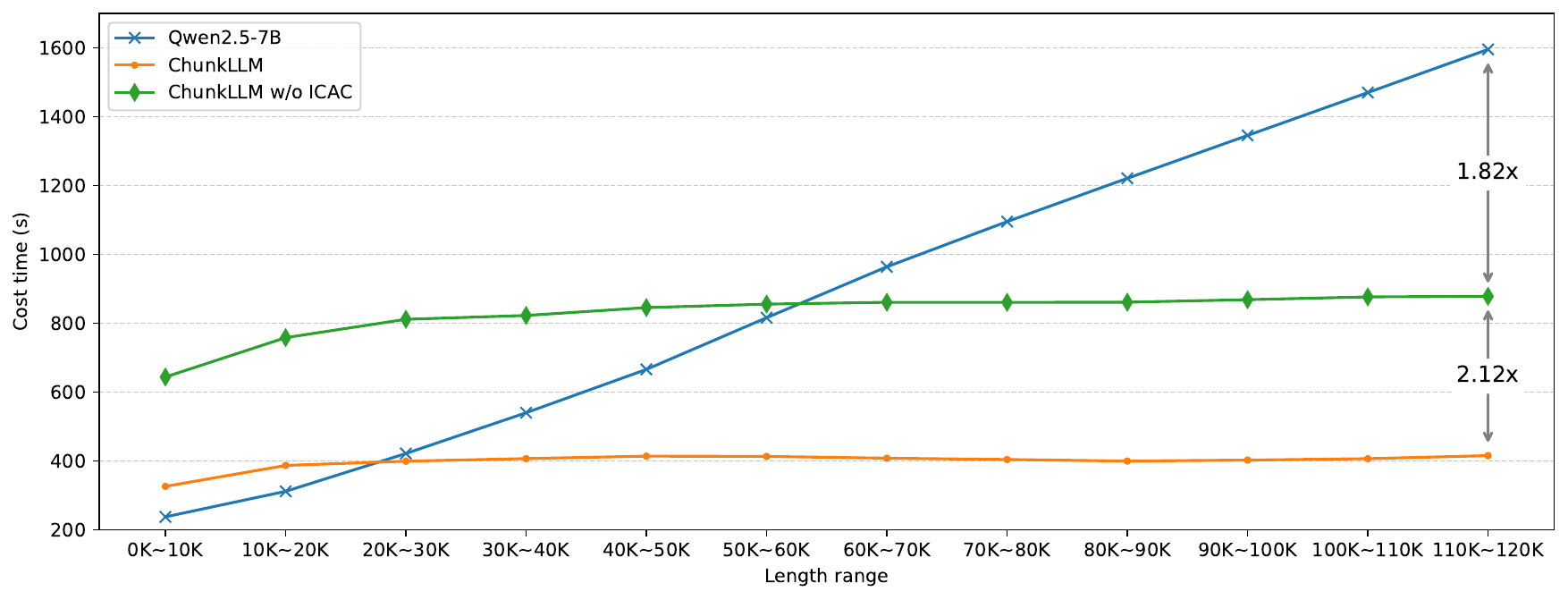}
\caption{Inference efficiency of ICAC on PG19 test set.}
\label{fig:efficiency_ablation}
\vspace{-0.5cm}
\end{figure}

\begin{figure*}[hbt]
    \centering
    \includegraphics[scale=0.4]{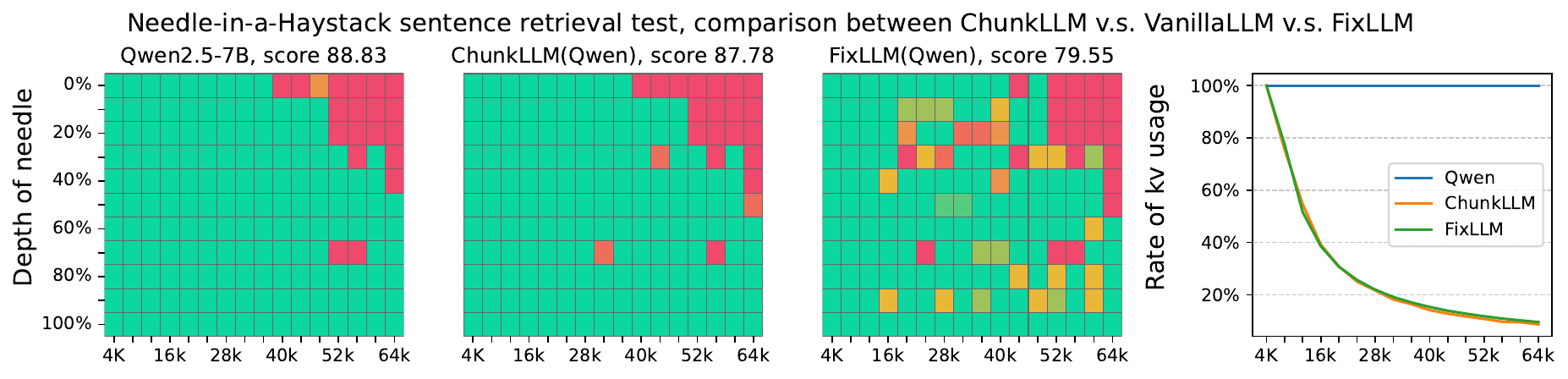}
    \caption{Visualization of fixed chunks and semantic chunks in NIAH test.
    }
    \label{fig:NIAH_fixed}
\end{figure*}

\subsection{Ablation Study}

\subsubsection{Effectiveness of Vote and ICAC}

We validate the proposed vote mechanism and ICAC pattern based on the Qwen2.5-7B using the LongBench, with experimental results shown in Table \ref{tab:longbench_ablation}. Removal of the vote mechanism leads to a 1.33 drop in overall performance (47.67 v.s. 46.34), which confirms the mechanism’s efficacy, as it integrates inter-layer differences in chunk selection and minimizes interference arising from such discrepancies.
We also conduct a visual investigation into the recall performance of top-k chunks across different layers, with comprehensive experimental results provided in Appendix \ref{top-k-chunks}.
Conversely, removing ICAC results in a marginal 0.09 improvement in overall performance. This slight gain is attributed to the increased frequency of chunk selection updates during the inference phase. 
Frequent chunk selection, however, poses a limitation of low inference efficiency. As shown in Figure \ref{fig:efficiency_ablation}, after ICAC is removed, the inference latency is 2.12 times higher than that of ChunkLLM in the 110K–120K token generation stage. Conversely, incorporating ICAC enables the model to maintain near-lossless performance alongside improved inference efficiency, which provides additional validation of ICAC’s success. Appendix \ref{case_study} shows a case study of the ICAC.

\subsubsection{Analysis of Fixed Chunks and Semantic Chunks}

We conduct an experimental analysis of the fixed chunk method (FixLLM) on the NIAH task. To ensure consistent KV cache utilization and facilitate a fair comparison, FixLLM is configured with a top-k of 384 and a local chunks of 24, while ChunkLLM is set to a top-k of 256 and a local chunks of 16. The experimental results are illustrated in Figure \ref{fig:NIAH_fixed}. As observed, under conditions of approximately consistent KV cache utilization, FixLLM exhibits an 8.23 reduction (87.78 v.s. 79.55) in accuracy relative to ChunkLLM on the 64K NIAH task. This discrepancy stems from the semantic incompleteness of fixed chunks, which in turn compromises chunk selection during the inference phase. In contrast, ChunkLLM leverages contextual semantic information to identify chunk boundaries, preserving the semantic integrity of chunks.

\section{Related Work}
\textbf{KV Cache Compression}
Recent research has primarily focused on overcoming the limitations of Large Language Models (LLMs) in processing massive contextual inputs.
SnapKV \citep{DBLP:conf/nips/LiHYVLYCLC24} improves efficiency through KV cache compression, using attention scores to select and cluster important positional information;
H2O \citep{DBLP:conf/nips/Zhang00CZC0TRBW23} implements a dynamic token retention policy that balances recent information and historically important information to optimize memory occupancy;
StreamingLLM \citep{DBLP:conf/iclr/XiaoTCHL24} enables LLMs to handle sequences of infinite length without fine-tuning by retaining attention sinks and local tokens;
PyramidInfer \citep{DBLP:conf/acl/YangHGHZ024} and PyramidKV \citep{DBLP:journals/corr/abs-2406-02069}  optimize performance by adjusting the KV cache capacity across different layers.
However, most methods in this category cannot be applied to the training phase.
MoBA \citep{DBLP:journals/corr/abs-2502-13189} proposes an innovative mixed block attention mechanism.
ESA \citep{DBLP:journals/corr/abs-2502-14477} reduces computational costs by selecting tokens most critical to the current generation for attention calculation. 

\textbf{Sparse Attention}
The sparse attention mechanism constructs sparse attention matrices by confining attention to predefined patterns, such as local windows or fixed-stride block patterns. \citet{DBLP:journals/corr/abs-2004-05150} combined dilated local window attention with task-specific global attention. 
NSA \citep{DBLP:conf/acl/YuanGD0ZZXWW0WR25} combines coarse-grained token compression and fine-grained token selection.
SepLLM \citep{DBLP:journals/corr/abs-2412-12094} finds that the segment information between separators can be effectively compressed into the separators themselves without causing significant information loss. 

\textbf{Knowledge Distillation}
Knowledge Distillation \citep{DBLP:journals/corr/HintonVD15}, as a widely used model compression technique, aims to train a student model under the guidance of a teacher model \citep{DBLP:journals/corr/RusuCGDKPMKH15,DBLP:journals/corr/abs-1910-01108,DBLP:journals/corr/abs-2006-05525}.
For text generation tasks, the standard KD method approximates the minimization of the forward Kullback-Leibler Divergence (forward KLD) between the generation distributions of the student and teacher models \citep{kim2016sequence,taori2023stanford,chiang2023vicuna,peng2023instruction,DBLP:journals/corr/abs-1910-01108}.

\section{Conclusion}

We introduce ChunkLLM, a lightweight and pluggable framework,
only necessitates fine-tuning lightweight components, QK Adapter and Chunk Adapter, on the basis of the original model architecture. 
And then we propose an attention distillation-based training approach for the QK Adapter, which leverages KL divergence to drive chunk attention toward approximating full attention, effectively enhancing the recall rate of key chunks. Furthermore, we introduce a novel "Intra-chunk Attention Consistency Pattern," which yields notable improvements in inference efficiency for long-context scenarios.
Experimental results show that ChunkLLM attains a maximum speedup of 4.48× in comparison to the vanilla Transformer in the processing of 120K long texts.

\section*{Limitations}

ChunkLLM primarily focuses on plain text tasks but is limited in handling complex types of textual tasks, such as tables or JSON-formatted data, which is not researcher-friendly.
In the future, we plan to incorporate table and JSON-formatted data into the training data to specifically enhance the model’s performance in these areas.





\bibliography{custom}

\appendix

\newpage
\section{Appendix}

\subsection{Parameter Setting and Training Datasets}
\label{paramer_set}
The Qwen2.5-7B and Llama3.1-8B models are trained with identical configurations. A cosine annealing strategy is adopted, with a maximum learning rate of 3e-5 and a warm-up period of 500 steps. Additionally, we use Adam optimizer with parameters beta1 = 0.9 and beta2 = 0.99.
For the Qwen2.5-7B and Llama3.1-8B, the dimensions of the QK Adapter and Chunk Adapter are set to 3584 and 4096, respectively. The total number of additional parameters is 14.7M and 21M, respectively.
$\alpha$ is set to 0.5 for chunk boundary prediction task. The training dataset comprised approximately 6B tokens, and the training process are conducted on 32 H200 GPUs for around 11,300 steps. We set training epoch is 1, and the attention distillation stage consum 11 hours, while the chunk boundary training stage took 1.5 hours. 

For a fair comparison on long-context benchmarks, we set the initial cache capacity to 4, the local cache capacity to 4000, separator cache capacity to 4000 and the maximum cache capacity to 8192 for SepLLM \citep{DBLP:journals/corr/abs-2412-12094}. For StreamingLLM \citep{DBLP:conf/iclr/XiaoTCHL24}, we configure the initial cache capacity as 4 and the local cache capacity as 8188. For SnapKV \citep{DBLP:conf/nips/LiHYVLYCLC24}, we set the cache capacity to 4096, the window size 32 and a max pooling operation with a kernel size of 7.

For short context benchmarks, we set the initial cache capacity to 4, the local cache capacity to 256, and the maximum cache capacity to infinity for SepLLM \citep{DBLP:journals/corr/abs-2412-12094}. For StreamingLLM \citep{DBLP:conf/iclr/XiaoTCHL24}, we configure the initial cache capacity as 4 and the local cache capacity as 256.

The FineWeb-Edu dataset \citep{lozhkov2024fineweb-edu} is employed as the training corpus in this study. Developed by the HuggingFaceFW team, this dataset undergoes filtering via an educational quality classifier, which constructed based on annotations generated by Llama3-70B-Instruct \citep{llama3modelcard}. 

For preprocessing the training data, the pySBD  tool \citep{sadvilkar-neumann-2020-pysbd}, a rule-based sentence boundary detection module that works out-of-the-box, is utilized to annotate the end positions of chunks in sequences, serving as foundational input for training the chunk boundary prediction module.

\subsection{Long Context Benchmarks}
\label{append_longbench}

The average text length for most tasks ranges from 5k to 15k tokens in LongBench. 
For NIAH, the benchmark constructs prompts for LLMs by randomly inserting key information into long texts. 

Table \ref{tab:append_longbench_subtask_description} presents the task name, abbreviations, evaluation methodologies, average lengths, and descriptive details of each subtask on LongBench.
\begin{table*}[t]
	\centering
    \scriptsize
	\begin{tabular}{lcccrp{5cm}}

        \hline
        Task & Subtask & Abbreviation & Evaluation & Avg Len & Description\\ 
        \hline
        \multirow{3}{*}{SDQA} & NarrativeQA & NQA & Recall &18,409 & Answer questions based on stories or scripts, including understanding of important elements such as characters, plots, themes, etc.\\
        \cdashline{2-6}[2pt/1pt]
        & Qasper & Qasper& Recall& 3,619& Answer questions based on a NLP research paper, questions proposed and answered by NLP practitioners.\\
        \cdashline{2-6}[2pt/1pt]
        & MultiFieldQA-en & MFQA & Recall  &4,559 & Answer English questions based on a long article, which comes from a relatively diverse field.\\
        \hline
        \multirow{3}{*}{MDQA} & HotpotQA &  HQA & Recall &9,151 & Answer related questions based on multiple given documents.\\
        \cdashline{2-6}[2pt/1pt]
        & Musique & Musi & Recall & 11,214 & Answer related questions based on multiple given documents.\\
        \cdashline{2-6}[2pt/1pt]
        & 2WikiMultihopQA & 2WQA & Recall & 4,887 & Answer related questions based on multiple given documents.\\
        \hline
        \multirow{2}{*}{Summary} & GovReport & GR & Rouge-L & 8,734 &A summarization task that requires summarizing government work reports.\\
        \cdashline{2-6}[2pt/1pt]
        & QMSum & QMS & Rouge-L & 10,614 & A summarization task that requires summarizing meeting records based on user queries.\\
        \hline
        \multirow{3}{*}{Few-shot} & SAMSum & SAM & Rouge-L &6,258& A dialogue summarization task, providing several few-shot examples.\\
        \cdashline{2-6}[2pt/1pt]
        & TriviaQA & TQA & F1 & 8209 & Single document question answering task, providing several few-shot examples. \\
        \cdashline{2-6}[2pt/1pt]
        & TREC & TREC & Accuracy& 5,177&A classification task that requires categorizing questions, includes 50 categories in total.\\
        \hline
        \multirow{2}{*}{Synthetic} & PassageCount & PC & Accuracy & 11,141 & Determine the total number of different paragraphs in a given repetitive article. \\
        \cdashline{2-6}[2pt/1pt]
        & PassageRetrieval-en & PR-en & Accuracy & 9,289 & Given 30 English Wikipedia paragraphs, determine which paragraph the given summary corresponds to. \\
	    \hline
        \multirow{2}{*}{Code} & LCC & LCC & Edit Sim & 1,235 & Given a long piece of code, predict the next line of code. \\
        \cdashline{2-6}[2pt/1pt]
        & RepoBench-P & RB-P & Edit Sim & 4,206 & Given code in multiple files within a GitHub repository (including cross-file dependencies), predict the next line of code. \\
        \hline

	\end{tabular}
	\caption{Task description on LongBench. SDQA: single-document question answering, MDQA: multi-document question answering.}
	\label{tab:append_longbench_subtask_description}
\end{table*}

\subsection{Top-K Sensitivity}
\label{top-k}

We set the local chunks to 15 and conduct experiments on LongBench to evaluate the hyperparameter top-k rate. The results are presented in Table \ref{tab:top-k}. Taking backbone Qwen2.5-7B as an example, the overall performance shows an upward trend as the top-k rate increases. This improvement is attributed to the introduction of more retrieved information, which enhances semantic richness. When the top-k rate is set to 45\% (ChunkLLM 47.67 v.s. Qwen2.5-7B 47.91) , the overall performance reaches 99.49\% of the vanilla model’s performance, demonstrating the effectiveness of our selected top-k rate. This observed pattern also holds for Llama3.1-8B.

\begin{table*}[h]
\setlength\tabcolsep{3pt} 
	\centering
    \scriptsize
	\begin{tabular}{llccccccccccccccc}
	    \hline
	    \multirow{2}{*}{Methods} & \multicolumn{3}{c}{SDQA} & \multicolumn{3}{c}{MDQA} & \multicolumn{2}{c}{Summary} & \multicolumn{3}{c}{Few-shot} & \multicolumn{2}{c}{Synthetic} & \multicolumn{2}{c}{Code} & \multirow{2}{*}{Avg}\\
        \cmidrule(lr){2-4} \cmidrule(lr){5-7} \cmidrule(lr){8-9}\cmidrule(lr){10-12} \cmidrule(lr){13-14} \cmidrule(lr){15-16}
        & NQA & Qasper & MFQA & HQA & Musi & 2WQA & GR & QMS & SAM & TQA & TREC & PC & PR-en & LCC & RB-P &\\
        \hline
        Qwen2.5-7B & 33.76 & 51.40 & 57.17 & 56.76 & 29.40 & 38.09 & 31.57 & 24.40 & 46.73 & 92.50 & 72.00 & 2.00 & 66.50 & 62.86 &53.53 &  47.91\\ 
        \cdashline{1-17}[2pt/1pt]
        ChunkLLM & 31.09 & 50.52 & 56.95 & 55.61 & 29.41 & 38.27 & 31.25 & 23.16 & 46.53 & 92.50 & 72.50 & 3.00 & 68.00 & 62.72 & 53.50 & 47.67\\
        Top-k rate 90 & 33.50 & 51.71 & 57.55 & 56.59 & 29.55 & 38.09 & 31.46 & 24.28 & 46.62 & 92.50 & 72.00 & 2.00 & 67.50 & 62.65 & 53.59 & 47.97\\
        Top-k rate 80 & 33.37 & 52.22 & 57.53 & 55.09 & 29.65 & 38.09 & 31.71 & 24.41 & 46.91 & 92.50 & 72.00 & 2.00 & 67.00 & 62.65 & 53.42 & 47.90 \\
        Top-k rate 70 & 32.27 & 52.31 & 57.63 & 55.09 & 30.01 & 38.43 & 31 59 & 23.89 & 46.97 & 92.50 & 72.00 & 2.50 & 65.50 & 62.64 & 53.74 & 47.80 \\
        Top-k rate 60 & 31.04 & 52.41 & 57.71 & 54.84 & 29.56 & 38.52 & 31.24 & 23.09 & 46.92 & 92.50 & 72.50 & 2.50 & 67.00 & 62.74 & 53.96 & 47.77 \\
        Top-k rate 50 & 30.40 & 50.76 & 58.64 & 54.78 & 29.41 & 37.67 & 31.70 & 22.97 & 46.86 & 92.50 & 72.50 & 3.00 & 67.50 & 62.80 & 53.58 & 47.68 \\
        Top-k rate 40 & 30.96 & 50.30 & 55.63 & 54.68 & 29.44 & 37.12 & 31.39 & 22.63 & 46.65 & 91.50 & 72.50 & 2.50 & 69.00 & 62.73 & 53.48 & 47.37 \\
        Top-k rate 30 & 30.21 & 48.70 & 54.08 & 54.65 & 29.37 & 37.27 & 31.35 & 22.45 & 46.79 & 91.50 & 72.50 & 2.50 & 67.00 & 62.66 & 53.25 & 46.95 \\
        Top-k rate 20 & 31.05 & 45.67 & 51.71 & 53.97 & 28.06 & 37.62 & 30.49 & 21.52 & 46.21 & 88.67 & 72.50 & 2.50 & 65.50 & 59.54 & 51.53 & 45.77 \\
        Top-k rate 10 & 30.74 & 41.09 & 48.42 & 53.80 & 26.17 & 36.07 & 29.52 & 21.19 & 45.64 & 84.92 & 72.00 & 1.50 & 64.00 & 57.20 & 49.14 & 44.09 \\
        \hline
        Llama3.1-8B & 40.88 & 51.72 & 58.47 & 52.68 & 35.80 & 42.52 & 30.64 & 24.44 & 47.10 & 92.83 & 73.00 & 0.50 & 39.00 & 68.56 & 67.48 &  48.37\\
        \cdashline{1-17}[2pt/1pt]
        ChunkLLM & 39.14 & 49.93 & 53.45 & 52.29 & 34.40 & 42.98 & 31.05 &  23.89 &   47.20 & 93.08 & 72.50 & 0.50 & 38.50 & 68.31 & 66.99 & 47.62\\
        Top-k rate 90 & 40.90 & 51.60 & 58.39 & 52.88 & 35.85 & 42.52 & 30.90 & 24.35 & 47.13 & 93.08 & 73.00 & 0.50 & 39.00 & 68.55 & 67.49 & 48.41 \\
        Top-k rate 80 & 40.91 & 51.92 & 57.57 & 52.33 & 35.77 & 42.85 & 31.30 & 24.45 & 47.19 & 92.83 & 73.00 & 0.50 & 39.00 & 68.55 & 67.57 & 48.38 \\
        Top-k rate 70 & 39.59 & 52.51 & 57.28 & 52.08 & 35.72 & 42.85 & 31.71 & 24.30 & 47.22 & 92.83 & 73.00 & 0.50 & 38.00 & 68.35 & 67.41 & 48.22 \\
        Top-k rate 60 & 40.18 & 51.18 & 55.26 & 51.71 & 35.78 & 42.85 & 31.82 & 23.72 & 46.44 & 93.08 & 73.00 & 0.50 & 39.00 & 68.32 & 67.51 & 47.99 \\
        Top-k rate 50 & 40.33 & 50.92 & 54.81 & 52.44 & 34.74 & 42.18 & 30.73 & 23.73 & 47.21 & 92.83 & 73.00 & 0.50 & 38.50 & 68.30 & 67.40 & 47.84 \\
        Top-k rate 40 & 37.88 & 48.12 & 53.21 & 52.58 & 33.71 & 41.98 & 30.00 & 23.92 & 47.25 & 93.08 & 73.00 & 0.50 & 37.50 & 68.24 & 67.12 & 47.21 \\
        Top-k rate 30 & 37.85 & 46.58 & 53.23 & 51.83 & 31.69 & 41.32 & 29.64 & 23.11 & 47.00 & 93.08 & 72.50 & 0.50 & 36.00 & 68.13 & 66.87 & 46.62 \\
        Top-k rate 20 & 37.16 & 44.96 & 52.19 & 51.32 & 29.96 & 40.68 & 26.49 & 22.94 & 46.14 & 92.83 & 72.50 & 0.50 & 35.50 & 67.15 & 66.10 & 45.76 \\
        Top-k rate 10 & 35.58 & 41.25 & 49.72 & 50.38 & 29.50 & 38.02 & 20.94 & 22.39 & 44.89 & 93.08 & 70.00 & 0.50 & 34.50 & 66.17 & 64.45 & 44.09 \\

	    \hline

	\end{tabular}
	\caption{Top-K sensitivity on LongBench.}
	\label{tab:top-k}
\end{table*}

\subsection{Local Chunks Sensitivity}
\label{local-chunks}

We also perform the hyperparameter experiment for local chunks on LongBench With the top-k rate fixed at 45\%. The experimental results are presented in Table \ref{tab:local-n}. Taking backbone Qwen2.5-7B as an example, the model’s overall performance improves as local chunks increase. This is because the number of local chunks directly affects the quality of the retrieved top-k chunks. More local chunks provide richer semantics, leading to the retrieval of more relevant chunks. This observed pattern also holds for Llama3.1-8B.

\begin{table*}[h]
\setlength\tabcolsep{3pt} 
	\centering
    \scriptsize
	\begin{tabular}{llccccccccccccccc}
	    \hline
	    \multirow{2}{*}{Methods} & \multicolumn{3}{c}{SDQA} & \multicolumn{3}{c}{MDQA} & \multicolumn{2}{c}{Summary} & \multicolumn{3}{c}{Few-shot} & \multicolumn{2}{c}{Synthetic} & \multicolumn{2}{c}{Code} & \multirow{2}{*}{Avg}\\
        \cmidrule(lr){2-4} \cmidrule(lr){5-7} \cmidrule(lr){8-9}\cmidrule(lr){10-12} \cmidrule(lr){13-14} \cmidrule(lr){15-16}
        & NQA & Qasper & MFQA & HQA & Musi & 2WQA & GR & QMS & SAM & TQA & TREC & PC & PR-en & LCC & RB-P &\\
        \hline
        Qwen2.5-7B & 33.76 & 51.40 & 57.17 & 56.76 & 29.40 & 38.09 & 31.57 & 24.40 & 46.73 & 92.50 & 72.00 & 2.00 & 66.50 & 62.86 &53.53 &  47.91\\ 
        \cdashline{1-17}[2pt/1pt]
        ChunkLLM & 31.09 & 50.52 & 56.95 & 55.61 & 29.41 & 38.27 & 31.25 & 23.16 & 46.53 & 92.50 & 72.50 & 3.00 & 68.00 & 62.72 & 53.50 & 47.67\\
        Local-n 30 & 31.63 & 51.03 & 57.48 & 55.78 & 29.10 & 38.68 & 31.91 & 22.85 & 46.86 & 92.00 & 72.50 & 2.50 & 69.00 & 62.64 & 53.63 & 47.84 \\
        Local-n 25 & 31.52 & 50.88 & 57.74 & 54.98 & 29.69 & 38.68 & 31.66 & 23.01 & 47.00 & 91.50 & 72.50 & 2.50 & 69.00 & 62.69 & 53.67 & 47.80\\
        Local-n 20 & 31.30 & 50.67 & 57.06 & 54.78 & 29.66 & 38.77 & 31.66 & 23.08 & 46.86 & 92.00 & 72.50 & 2.50 & 68.00 & 62.68 & 53.46 & 47.67 \\
        Local-n 10 & 30.44 & 51.16 & 57.35 & 54.68 & 29.66 & 38.27 & 31.51 & 22.91 & 46.39 & 91.50 & 72.50 & 3.00 & 68.00 & 62.72 & 53.48 & 47.57 \\
        Local-n 5  & 29.71 & 50.23 & 56.22 & 54.44 & 29.36 & 37.77 & 31.51 & 22.99 & 46.13 & 91.50 & 72.50 & 2.50 & 68.00 & 62.40 & 52.70 & 47.17 \\
        Local-n 0  & 29.74 & 49.01 & 55.16 & 54.28 & 29.16 & 37.77 & 31.74 & 23.16 & 46.35 & 91.50 & 72.50 & 2.50 & 68.00 & 35.01 & 44.20 & 44.67 \\
        
        \hline
        Llama3.1-8B & 40.88 & 51.72 & 58.47 & 52.68 & 35.80 & 42.52 & 30.64 & 24.44 & 47.10 & 92.83 & 73.00 & 0.50 & 39.00 & 68.56 & 67.48 &  48.37\\
        \cdashline{1-17}[2pt/1pt]
        ChunkLLM & 39.14 & 49.93 & 53.45 & 52.29 & 34.40 & 42.98 & 31.05 &  23.89 &   47.20 & 93.08 & 72.50 & 0.50 & 38.50 & 68.31 & 66.99 & 47.62\\
        
        Local-n 30 & 39.44 & 50.53 & 54.97 & 52.53 & 34.97 & 42.58 & 31.02 & 24.07 & 46.97 & 93.08 & 73.00 & 0.50 & 39.00 & 68.49 & 67.27 & 47.89 \\
        Local-n 25 & 39.63 & 50.39 & 54.24 & 52.53 & 34.72 & 42.43 & 31.17 & 23.96 & 46.87 & 93.08 & 73.00 & 0.50 & 38.50 & 68.47 & 67.16 & 47.78 \\
        Local-n 20 & 39.33 & 49.98 & 53.71 & 52.38 & 34.50 & 42.75 & 30.98 & 23.77 & 47.05 & 93.08 & 72.50 & 0.50 & 38.50 & 68.48 & 67.12 & 47.64 \\
        Local-n 10 & 39.06 & 48.71 & 52.67 & 51.58 & 34.35 & 42.48 & 30.79 & 24.16 & 46.87 & 92.83 & 72.50 & 0.50 & 37.50 & 68.15 & 66.69 & 47.26 \\
        Local-n 5  & 39.02 & 48.66 & 52.67 & 51.54 & 34.24 & 42.48 & 30.30 & 24.14 & 46.58 & 93.08 & 72.50 & 0.50 & 37.00 & 67.22 & 66.61 & 47.10 \\
        Local-n 0  & 38.91 & 48.39 & 52.03 & 51.58 & 34.24 & 42.02 & 30.41 & 23.77 & 45.99 & 93.08 & 72.50 & 0.50 & 37.00 & 36.57 & 55.18 & 44.14 \\

	    \hline

	\end{tabular}
	\caption{Local chunks sensitivity on LongBench.}
	\label{tab:local-n}
\end{table*}

\subsection{Compare with StreamingLLM on NIAH}
\label{NIAH_stream}

\begin{figure*}[t]
    \centering
    \includegraphics[scale=0.4]{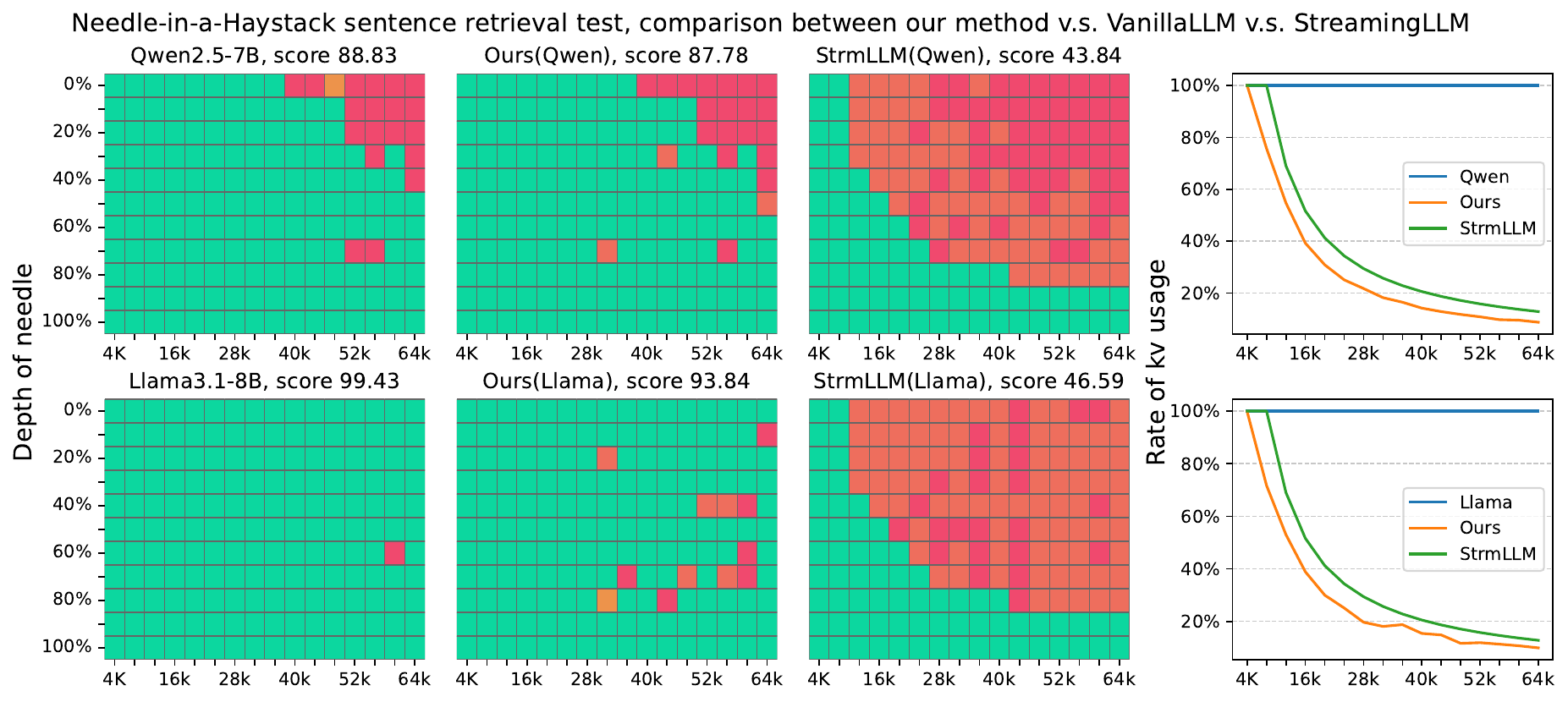}
    \caption{Compare with StreamingLLM on Needle-in-a-Haystack. The last column represents the KV-cache utilization rate.
    }
    \label{fig:NIAH_stream}
\end{figure*}

As depicted in Figure \ref{fig:NIAH_stream}, ChunkLLM outperforms StreamingLLM across all scenarios in the 64K-context NIAH evaluation conducted on Qwen2.5-7B and Llama3.1-8B with lower KV-cache usage, achieving superior performance. 
SteamingLLM leverages initial tokens and local tokens as its core token selection strategy. However, this design inherently limits its ability to effectively capture critical information situated in the middle segment of the input sequence, thereby resulting in a notable deficiency in long-context retrieval performance.

\subsection{Inference Time of All Sub-processes}
\label{sub-process}

\begin{figure*}[hbt]
    \centering
    \begin{subfigure}{\textwidth}
        \includegraphics[width=\linewidth]{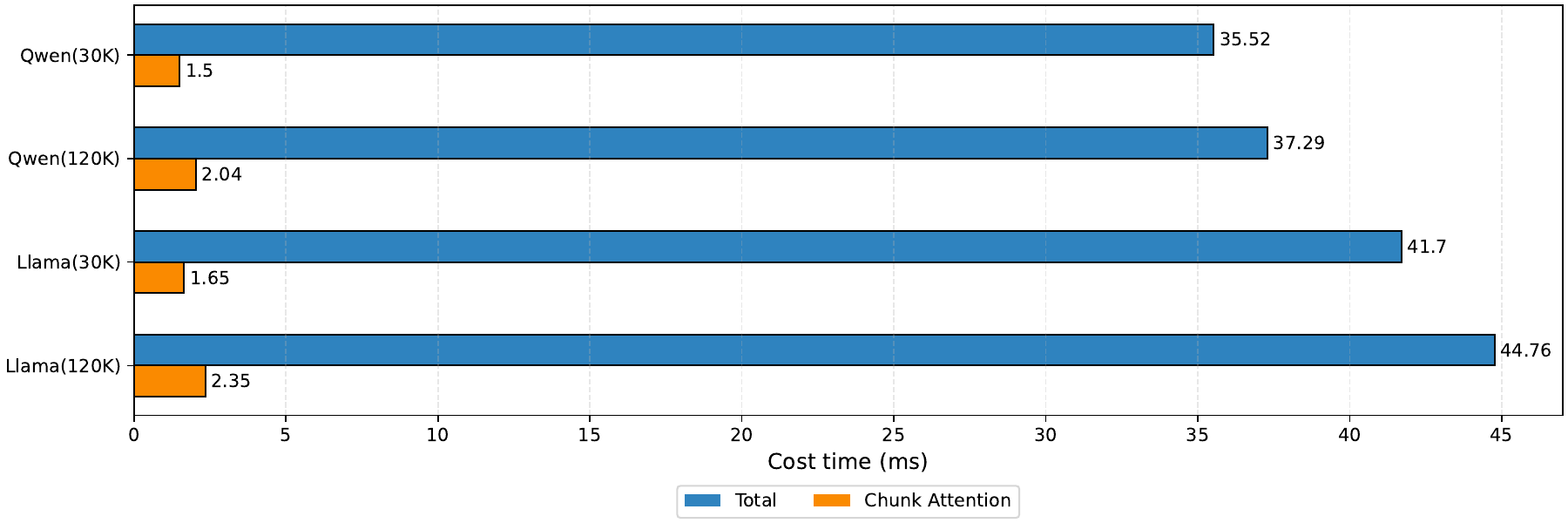}
        \caption{Total time and time consumed by the chunk attention mechanism during inference.}
    \end{subfigure}

    \begin{subfigure}{\textwidth}
        \includegraphics[width=\linewidth]{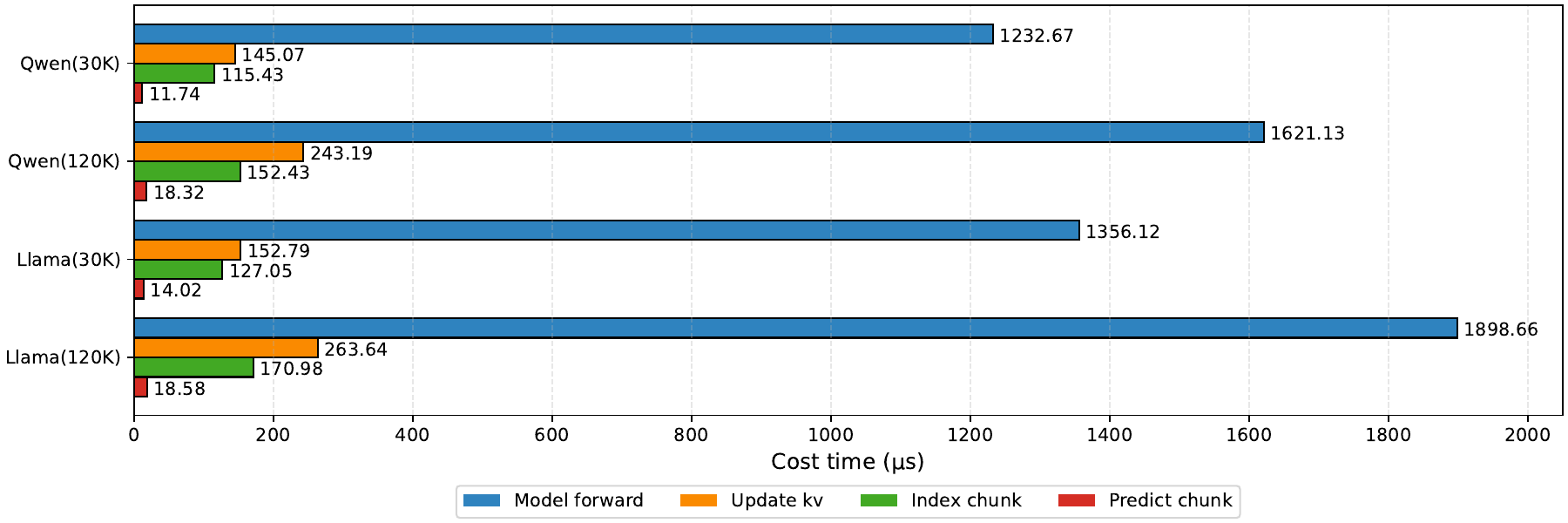}
        \caption{Cost time of each sub-process in chunk attention  mechanism.}
    \end{subfigure}

    \begin{subfigure}{\textwidth}
        \includegraphics[width=\linewidth]{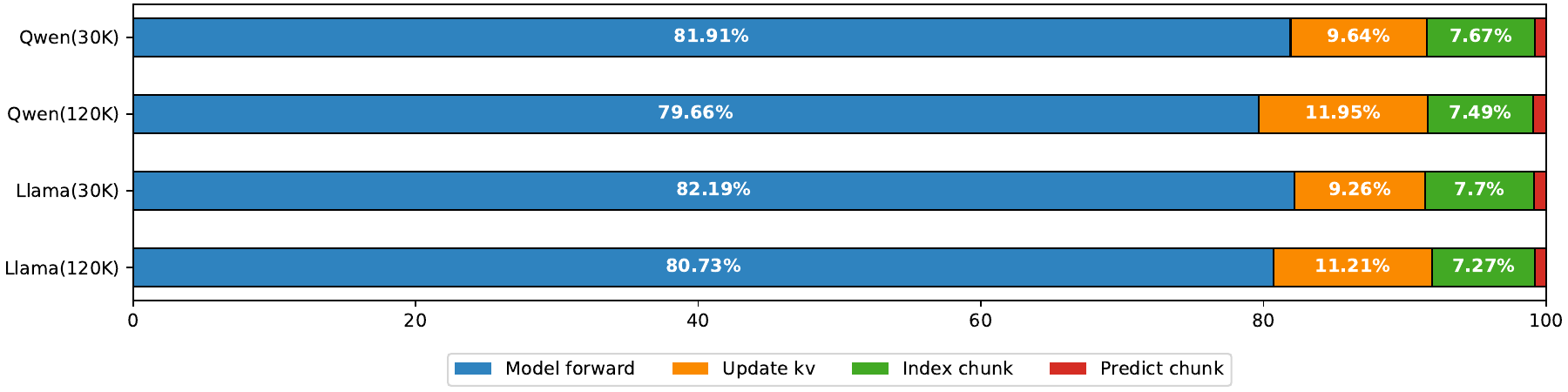}
        \caption{Time proportion of each sub-process in chunk attention  mechanism.}
    \end{subfigure}

    \caption{Inference time of all sub-processes.}
    \label{fig:inference-all}
\end{figure*}

We measure the per-token total average time and chunk attention mechanism time for ChunkLLM with varying input lengths on PG19 test set during decoding, using Qwen2.5-7B and Llama3.1-8B backbones. (Times are averaged over 1000 generated tokens.)
As shown in Figure \ref{fig:inference-all}(a), at an input length of 120K tokens, the chunk attention mechanism contributes only about 5.47\% (2.04 v.s. 37.29) and 5.25\% (2.35 v.s. 44.76) to the total time, respectively. This result demonstrates its low overhead.

In particular, we further analyze the time consumption of individual sub-processes within the chunk attention mechanism, which consists of four parts: model forward, KV cache update,chunk retrieval (Index chunk), and chunk boundary prediction (Predict chunk). The respective time costs and their proportions are illustrated in Figure \ref{fig:inference-all}(b) and (c).

It is evident that the model forward pass consumes around 80\% of the time, whereas chunk retrieval(Index Chunk) and boundary prediction(Predict chunk) require a negligible 7\% and 1\%, respectively. The minimal cost of these auxiliary operations validates the high inference efficiency of ChunkLLM.

\subsection{Case Study}
\label{case_study}

\begin{figure*}[h]
    \centering
    \begin{subfigure}{\textwidth}
        \includegraphics[width=\linewidth]{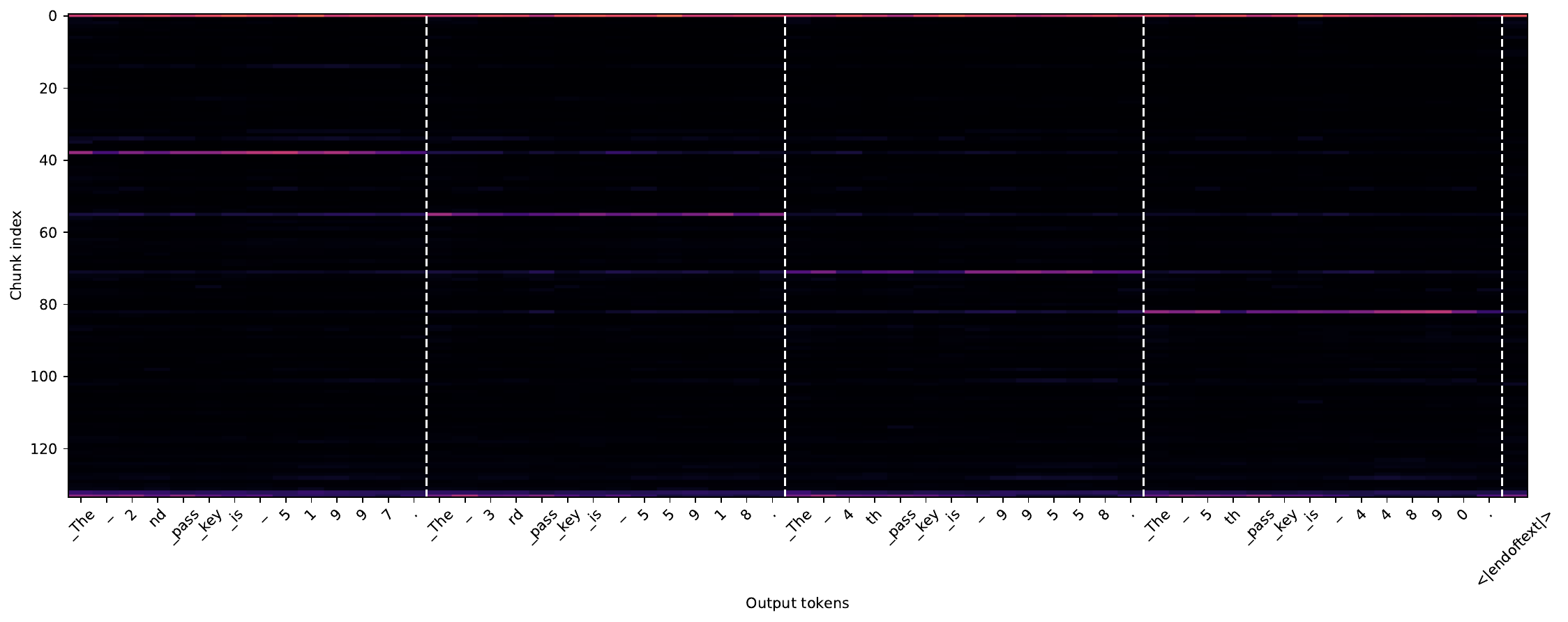}
        \caption{The sample is derived from the passkey retrieval task.}
    \end{subfigure}

    \begin{subfigure}{\textwidth}
        \includegraphics[width=\linewidth]{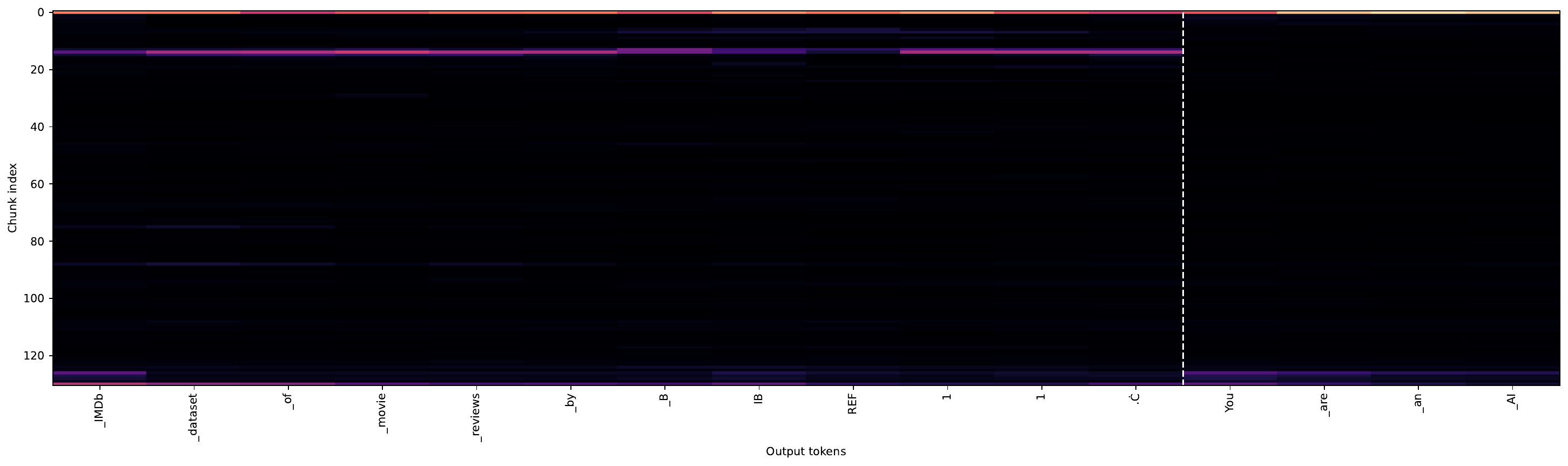}
        \caption{The sample is derived from the Qasper task in LongBench.}
    \end{subfigure}

    \begin{subfigure}{\textwidth}
        \includegraphics[width=\linewidth]{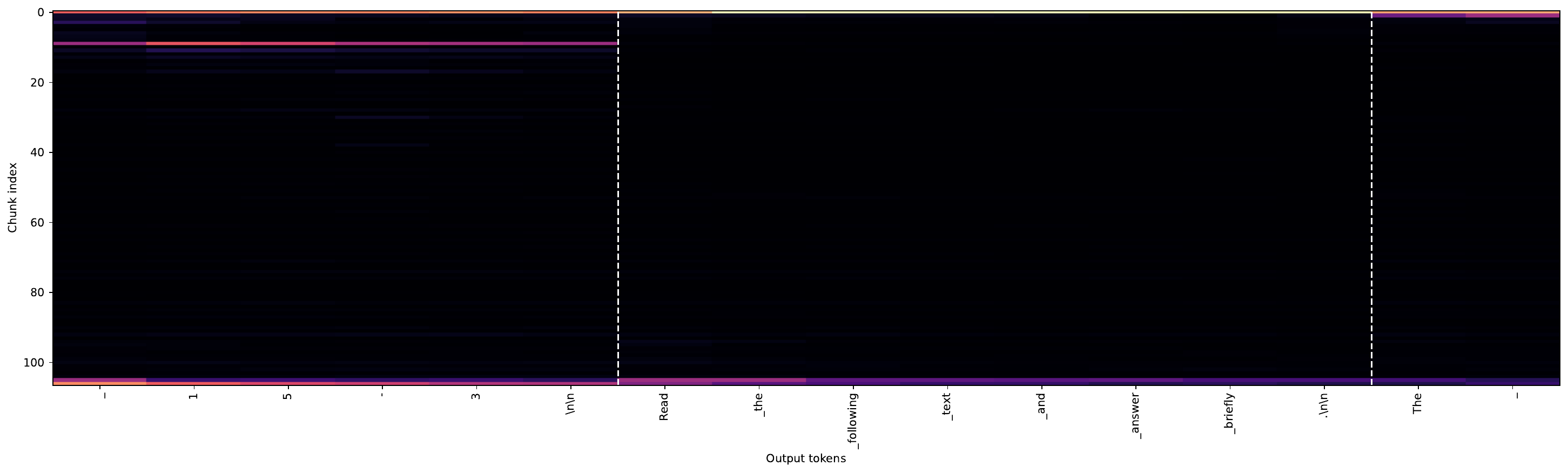}
        \caption{The sample is derived from the MultiFieldQA-en task in LongBench.}
    \end{subfigure}

    \caption{Case study of ICAC.}
    \label{fig:case_study}
\end{figure*}

To illustrate the reasoning process, we randomly sample three examples from the passkey retrieval, Qasper and MultiFieldQA-en tasks in LongBench, with its visualization results presented in Figure \ref{fig:case_study}. As observed in the figure, during the generation phase, the chunks attended to by tokens within the same chunk demonstrate remarkably high consistency, corresponding to the highlighted bands in the visualization, whereas shifts in attention occur exclusively at chunk boundaries. This empirical observation validates the effectiveness of the proposed ICAC pattern.

\subsection{Top-k Chunks Cross All Layers}
\label{top-k-chunks}
\begin{figure*}[hbt]
    \centering
    \begin{subfigure}{0.7\textwidth}
        \includegraphics[width=\linewidth]{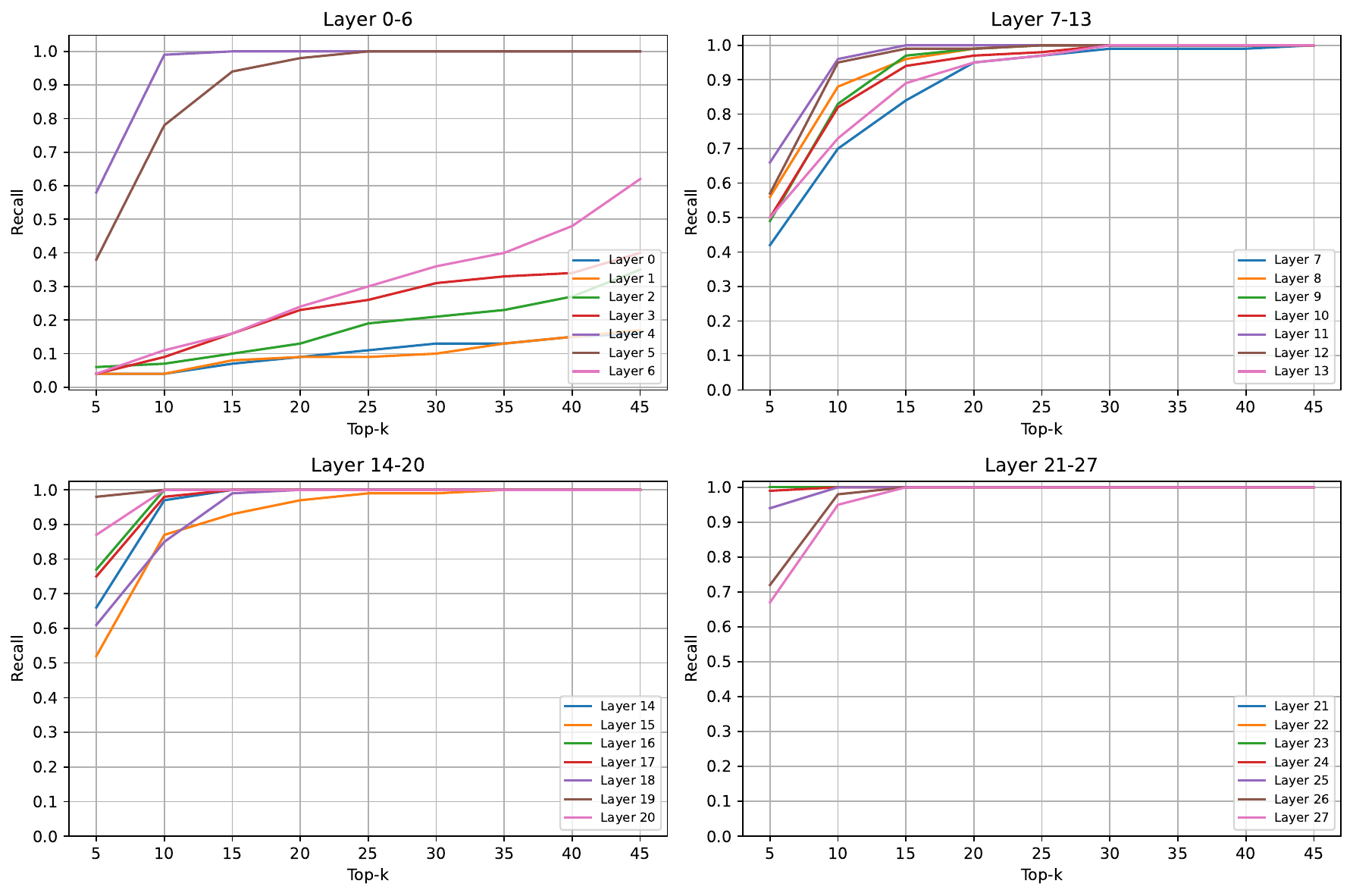}
        \caption{Qwen2.5-7B with 4K input length.}
    \end{subfigure}

    \begin{subfigure}{0.7\textwidth}
        \includegraphics[width=\linewidth]{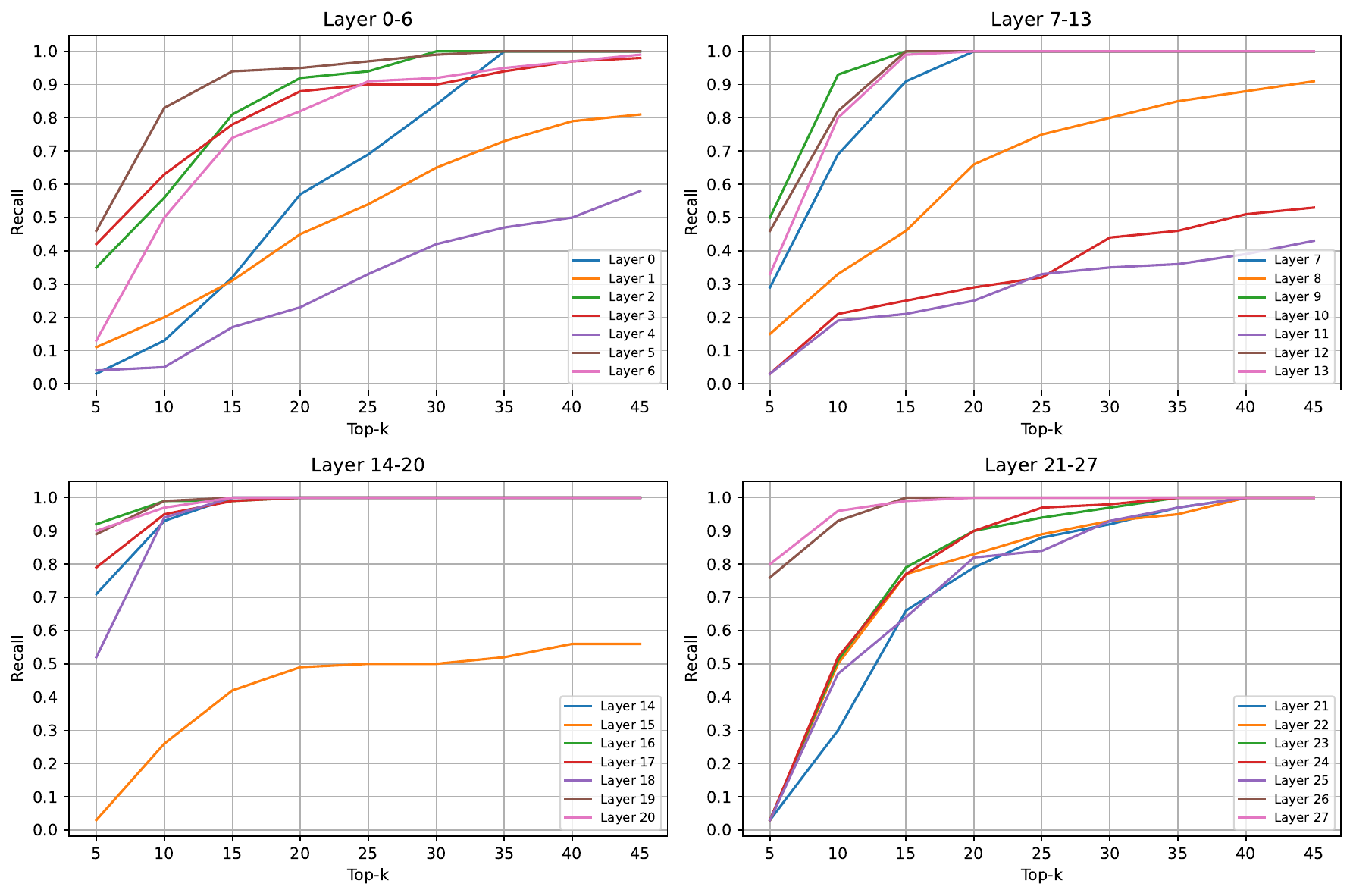}
        \caption{ChunkLLM with 4K input length.}
    \end{subfigure}

    \begin{subfigure}{0.7\textwidth}
        \includegraphics[width=\linewidth]{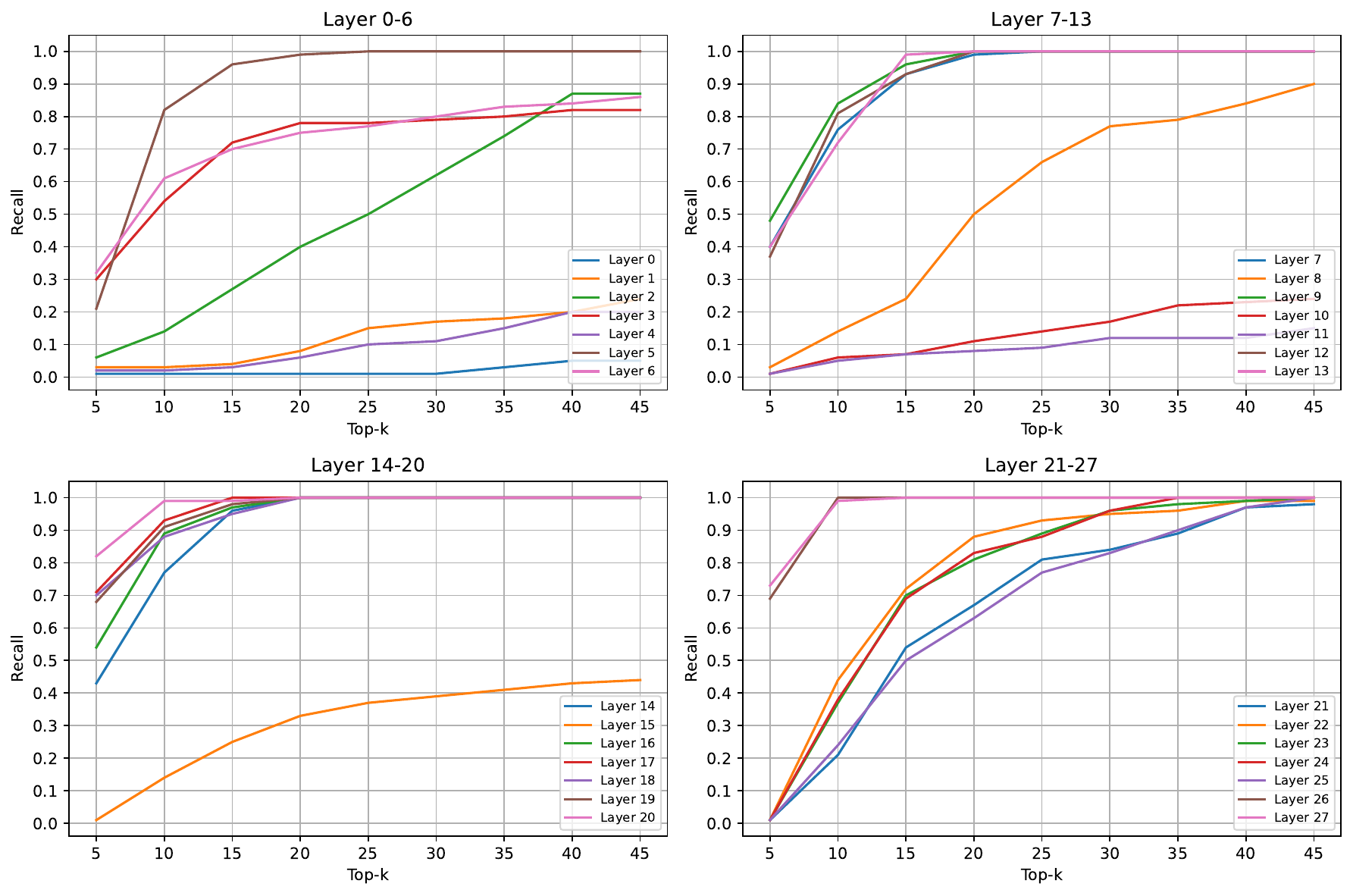}
        \caption{ChunkLLM with 32K input length.}
    \end{subfigure}

    \caption{Top-k chunks aross all layers.}
    \label{fig:top_k_chunks}
\end{figure*}

Figure \ref{fig:top_k_chunks} presents the recall performance of top-k chunks across all layers based on Qwen2.5-7B. A consistent pattern emerges: the chunk recall rate in the lower layers (Layers 0–6) is relatively low, which we attribute to insufficient semantic representation. In contrast, the middle layers (Layers 7–20) demonstrate a notably higher chunk recall rate. Specifically, when top-k is set to 15, the recall rate of these middle layers exceeds 80\%. 
This phenomenon, we contend, stems from the richer semantic representations inherent in the middle layers, coupled with the fact that our proposed attention distillation strategy effectively enhances the model’s chunk selection capability. 
Conversely, the chunk recall rate in the highest layers (Layers 21–27) exhibits a downward trend; we attribute this to the functional role of the highest layers, which are primarily dedicated to facilitating the model’s output generation. Notably, the chunk voting mechanism can effectively mitigate discrepancies between cross-layer chunks, 
and thereby enables the achievement of optimal performance.

\end{document}